\documentclass[1p]{elsarticle}

\usepackage[utf8]{inputenc}
\usepackage{graphicx}
\usepackage{algorithm , algorithmic}
\usepackage{varwidth}
\usepackage[english]{babel}
\usepackage{amssymb, amsthm, amsmath}
\usepackage{amssymb,amsmath}
\usepackage{array}
\usepackage{rotating}
\usepackage{multirow}
\usepackage{stmaryrd}
\usepackage{booktabs}
\usepackage[font=small,labelfont=bf]{caption}
\usepackage{subcaption}
\usepackage{color}
\usepackage{epsfig}
\usepackage{colortbl}
\usepackage[table]{xcolor}
\usepackage{hhline}
\usepackage[autolanguage]{numprint}
\usepackage{mattens}
\usepackage{natbib}
\usepackage{hyperref} %\usepackage[draft]{hyperref}
\usepackage{float, lscape}
\theoremstyle{definition}
\newtheorem{definition}{Definition}
\newtheorem{proposition}{Proposition}
\newtheorem{theorem}{Theorem}

\newcommand{\AMOCO}{\emph{TPA}}

\begin{document}

%Original
%\title{Effective anytime algorithm for multiobjective combinatorial optimization problems}
%\author{Miguel Ángel Domínguez-Ríos, Francisco Chicano, Enrique Alba}
%\maketitle
%
 
%Preprint
\begin{frontmatter}
\title{Effective anytime algorithm for multiobjective combinatorial optimization problems}
\author[1]{Miguel {\'A}ngel Dom{\'i}nguez-R{\'i}os\corref{cor1}}
\ead{miguel.angel.dominguez.rios@uma.es}
\author[1]{Francisco Chicano}
\ead{chicano@lcc.uma.es}
\author[1]{Enrique Alba}
\ead{eat@lcc.uma.es}
%\address[1]{Dept. Lenguajes y Ciencias de la Computación. Universidad de Málaga (Spain)}
\address[1]{ITIS Software. Universidad de M{\'a}laga (Spain)}
\cortext[cor1]{Corresponding author}

\hyphenation{me-tric}

\begin{abstract}
In multiobjective optimization, the result of an optimization algorithm is a set of efficient solutions from which the decision maker selects one. It is common that not all the efficient solutions can be computed in a short time and the search algorithm has to be stopped prematurely to analyze the solutions found so far. 
A set of efficient solutions that are well-spread in the objective space is preferred to provide the decision maker with a great variety of solutions. However, just a few exact algorithms in the literature exist with the ability to provide such a well-spread set of solutions at any moment: we call them \emph{anytime} algorithms. We propose a new exact anytime algorithm for multiobjective combinatorial optimization combining three novel ideas to enhance the  anytime behavior. We compare the proposed algorithm with those in the state-of-the-art for anytime multiobjective combinatorial optimization using a set of 480 instances from different well-known benchmarks and four different performance measures: the overall non-dominated vector generation ratio, the hypervolume, the general spread and the additive epsilon indicator. A comprehensive experimental study reveals that our proposal outperforms the previous algorithms in most of the instances.
\end{abstract}

%Preprint
\end{frontmatter}

\section{Introduction} \label{sec:introduction} 
MultiObjective Optimization (MOO) is a field of research with many applications in different areas, such as biology, computer science, scheduling, and finances. Broadly speaking, in a multiobjective optimization problem some objectives are in conflict and they have to be optimized simultaneously. Without loss of generality, we consider minimization problems throughout this paper. In the case of a maximization problem, we use the property $max(f)=-min(-f)$. 
A MultiObjective Program (MOP) is an optimization problem characterized by multiple and conflicting objective functions that are to be optimized over a feasible set of decisions. 
If the constraints and the objective functions are linear, the MOP is a MultiObjective Linear Program (MOLP), and when the variables are integer, we name it MultiObjective Integer Program (MOIP). 
If some variables are constrained to be integer-valued and the rest are continuous, the problem is a MultiObjective Mixed Integer Program (MOMIP).
MultiObjective Combinatorial Optimization (MOCO) is a special case of MOIP when the feasible set is finite. MOIP and MOCO are special cases of MultiObjective Discrete Optimization (MODO).

Finding the whole Pareto front in MOCO problems can require too much computational time when a faster solution is needed. For example, one may need to solve an assignment problem in hours and finding the whole Pareto front can take days. In particular, this happens in instances with many non-dominated points, or when the model is hard in itself. From a practical point of view, the decision maker with whom we interact may be interested in a set of solutions as well-spread as possible in the objective space at any time during the search. 
The concept of \emph{anytime} algorithms was introduced by Dean and Boddy~\cite{dean1988solving} to characterize algorithms with the following two properties: \emph{(i)} they can be terminated at any time and still return an answer, and \emph{(ii)} the answers returned improve in some well-behaved manner as a function of time. In a more precise way, algorithms that show a better trade-off between solution quality and runtime are said to have a better anytime behavior~\cite{lopez2014automatically}.

In this paper we propose a new exact algorithm to solve MOCO problems, which is also valid for MODO problems with finite feasible sets. 
The fact that the algorithm is exact and all the solutions found are in the Pareto front allows less freedom when interpreting the algorithm and reduces the probability that two different implementations behave differently~\cite{rostami2020algorithmic}. 
The algorithm is anytime in the sense that it is possible to interrupt its execution and take the non-dominated set. Moreover, it is guaranteed that this set is well-spread in the objective space. To check this spread and the quality of the solutions properly, four well-known metrics in the literature have been used: the overall non-dominated vector generation ratio~\cite{jiang2014consistencies}, the hypervolume indicator~\cite{fonseca2006improved, while2012fast}, the general spread~\cite{jiang2014consistencies, zhou2006combining}, and the additive epsilon indicator~\cite{liefooghe2016correlation}. In the computational results section, all these metrics are calculated for each of the 480 instances considered in the work presented here.

The proposed algorithm is based on a general framework by Dächert and Klamroth~\cite{dachert2015linear}. Broadly speaking, it consists in analyzing a search region in the objective space and looking for new non-dominated points at each iteration. These regions can be considered as boxes in $\mathbb{R}^{p}$, where $p$ is the number of objectives. The contribution of this paper is to present new designing criteria and an innovative way of splitting the objective space so that the solutions obtained are well-spread over the objective space. These are detailed in the following three novel proposals:

\begin{itemize}
    \item  A new strategy to select the appropriate search space region as the next box to explore, in order to guarantee that the new solutions found are spread over the objective space.
    \item A new way of partitioning the search space after finding a new non-dominated point. This partition reorders the selection of the future boxes to explore and also has an influence in the spread of solutions.
    \item A new quality function to measure the priority for the new regions to explore.
\end{itemize}

In order to help readers to reproduce the results of this paper, we uploaded the source code to GitHub\footnote{Available at \url{https://github.com/MiguelAngelDominguezRios/boxMO}.}. This is a good practice that is becoming popular to face the reproducibility crisis in Computer Science~\cite{Hutson725}.

The paper is organized as follows. In Section \ref{sec:literature_review}, we provide a literature review including the most relevant papers on MOO algorithms that could be used as anytime MOCO algorithms with good spread. In Section~\ref{sec:background}, we provide the background with some general definitions and the metrics we use in the computational results. In Section~\ref{sec:general-formulation}, a general formulation to solve MOCO problems is given. 
This formulation is the backbone of our contribution. 
In Section~\ref{sec:new_algorithm} our new anytime algorithm is presented. An exhaustive computational analysis is shown in Section~\ref{sec:computational_results}. The last section is dedicated to the conclusions and future work.

\section{Literature review} \label{sec:literature_review}
In this section we review different exact algorithms to solve multiobjective problems. We organize this review in four subsections. The first three subsections cover related works, which are presented (mostly) in chronological order in each subsection. The subsections focus respectively on the first algorithms, the ones based on recursion, and more recent algorithms. The fourth subsection summarizes the algorithms which can be considered as anytime with good spread and that are used to compare with our proposed algorithm.

\subsection{First exact algorithms}
In 2004, Sylva and Crema~\cite{sylva2004method}, working on the idea of Klein et al.~\cite{klein1982algorithm}, designed a new method to solve MultiObjective Integer Linear Programming (MOILP). They partitioned the solution space of dimension $p$ by adding  $p$ binary variables and $p$ constraints to exclude the previously-generated non-dominated points. About three years later, Sylva and Crema~\cite{sylva2007method} formulated a variant of their previous work, finding well-spread subsets of non-dominated points in MOMIP. If the variables are integer, the whole Pareto front is obtained.
To the best of the authors' knowledge, this was the first work to show dispersed solutions in the objective space, supported by computational results. 

Masin and Bukchin~\cite{masin2008diversity} developed an algorithm for MOMIP, called DMA (Diversity Maximization Approach), which finds solutions by maximizing a proposed diversity measure and guarantees the generation of the complete set of efficient points. 

In 2013, Lokman and Köksalan~\cite{lokman2013finding} proposed two improved algorithms based on the work of Sylva and Crema~\cite{sylva2004method}. The algorithms iteratively generate non-dominated points and exclude the regions that are dominated by the previously-generated non-dominated points. 

Ceyhan et al.~\cite{ceyhan2019finding}, based on the work of Lokman and Köksalan~\cite{lokman2013finding}, proposed some algorithms to find a small representative set of non-dominated points for MOMIP. The first of them, called SBA, is able to find the complete Pareto front for discrete problems if enough time is available. The main advantage of the SBA algorithm is that it produces very well-spread solutions over the objective space. 
As noted in this paper~\cite{ceyhan2019finding}, the work of Masin and Bukchin~\cite{masin2008diversity} is similar to Sylva and Crema's method in~\cite{sylva2007method}, because they generate similar representative subsets of non-dominated points for MOMIPs. In fact, SBA outperformed Masin and Bukchin's algorithm.

\subsection{Algorithms based on recursion}

In 2003, Tenfelde-Podehl~\cite{tenfelde2003recursive} proposed a recursive algorithm for MOCO problems with $p$ criteria. Dhaenens et al.~\cite{dhaenens2010k} proposed a new method to solve MOP, called $K$-PPM (Parallel Partitioning Method), which is based on the work of Lemesre et al.~\cite{lemesre2007parallel}, called 2-PPM and valid only for the bi-objective case. This new approach is valid for any dimension $p \geq 2$, and could also be considered as an extension of the idea of Tenfelde-Podehl~\cite{tenfelde2003recursive}. 

Özlen and Azizo{\u{g}}lu~\cite{ozlen2009multi} developed a general approach to generate the Pareto front for MOIP problems, based in the $\varepsilon$-constraint method. They also used recursion to solve multiobjective problems with fewer objectives obtaining efficiency ranges for each objective.  
One drawback of the method in~\cite{ozlen2009multi} is that it generates the same non-dominated point many times. The algorithm was enhanced in the work of Özlen et al.~\cite{ozlen2014multi}.

\subsection{Modern algorithms}
Laumanns et al.~\cite{laumanns2006efficient, laumanns2005adaptive} proposed the first algorithm to calculate the Pareto front in MOP through the resolution of a number of single-objective problems that depend only on the number of solutions in the front. The work of Laumanns et al.~\cite{laumanns2006efficient} was based on the  $\varepsilon$-constraint method. 
They obtained a bound $O(k^{p-1})$ for the number of calls to the single-objective solver, where $k$ is the number of non-dominated points and $p$ the number of objectives.
In~\cite{laumanns2005adaptive}, Laumanns et al. presented another algorithm for solving MOP using only lower bounds (for maximization problems) which leads to a fewer number of constraints. 
The algorithm by Kirlik and Say\i n~\cite{kirlik2014new} was an improvement of the work of Laumanns et al.~\cite{laumanns2006efficient}. They changed the order in which the subproblems were solved. The search is managed over ($p-1$)-dimensional rectangles.

In the last decade, modern algorithms solving MOO problems have been developed. Apart from the aforementioned work by Ceyhan et al.~\cite{ceyhan2019finding}, there are other important papers we should mention. Przybylski et al.~\cite{przybylski2010recursive} proposed an algorithm to compute the supported non-dominated extreme points for MOIP. 
This work was generalized by Przybylski et al.~\cite{przybylski2010two} to obtain the complete Pareto front for MOIP and is considered as a generalization of the two-phase method of Ulungu and Teghem~\cite{ulungu1995two}. 
\"Ozpeynirci and Köksalan~\cite{ozpeynirci2010exact}, among other authors, developed an algorithm to generate supported non-dominated points for MOMIP.
Dächert and Klamroth~\cite{dachert2015linear} conceived an algorithm to solve MOO for $p=3$. Their method is based on the work of Przybylski et al.~\cite{przybylski2010recursive} and~\cite{przybylski2010two}. They named \emph{full m-split} the general framework for solving any MOO of dimension $m$. This algorithm, suitably adapted, is used in our work. 
Later, Klamroth et al.~\cite{klamroth2015representation} improved \emph{full m-split}, developing two new methods, called $RE$ and $RA$, where they were concerned with eliminating or avoiding redundancies in the search zone explorations.
Recently, Dächert et al.~\cite{dachert2017efficient} have provided new theoretical insights into structural properties of the search region, assuming that a finite set of mutually non-dominated points has already been computed. The work provides better results than those in~\cite{klamroth2015representation}  for high dimensions ($p\geq6$). In 2018, Holzmann and Smith~\cite{holzmann2018solving} designed an algorithm to solve MODO problems using a weighted augmented Tchebycheff scalarization. This work is also based on the work of Dächert and Klamroth~\cite{dachert2015linear}. They improved the results of Kirlik and Say{\i}n~\cite{kirlik2014new} and Özlen et al.~\cite{ozlen2014multi}.

\subsection{Related algorithms that can be considered as anytime with good spread} 
\label{sec:related_algorithms}

As stated in the preceding review, only a few methods can be considered to have good spread over the objective space. This is in concordance with the work of Ceyhan et al.~\cite{ceyhan2019finding}, which summarized in two the number of algorithms which can be used to calculate the complete front in MOCO problems~\cite{masin2008diversity,sylva2007method}. If fact, with their SBA algorithm, they outperformed the results in~\cite{sylva2007method}. 
The work of Holzmann and Smith~\cite{holzmann2018solving} can also be considered as anytime, so we include it in our study.

We analyze which other algorithms in the literature can be slightly modified to obtain a well-spread set of non-dominated points at any time during the search.
The algorithms which use the $\varepsilon$-constraint method do not seem to be good at spreading solutions over the objective space, because in the objective function, just one selected objective is optimized (Ch. 4 of~\cite{ehrgott2005multicriteria}). Moreover, methods which use branch and bound as the general framework to obtain the Pareto front are widely surpassed by modern algorithms, which avoid repeating solutions~\cite{dachert2015linear}. In addition, those which calculate the nadir point in a first phase can be slow because computing the nadir point is an NP-hard problem in  itself~\cite{gary1979computers}, so we do not consider these algorithms.

In conclusion, we have chosen two algorithms: SBA, from the work of Ceyhan et al.~\cite{ceyhan2019finding}, and the method of Holzmann and Smith~\cite{holzmann2018solving}. They are, to the best of the authors' knowledge, the algorithms that can potentially provide a well-spread set of non-dominated points at any time of execution. 

\section{Background} \label{sec:background}
In this section, we first present some basic definitions to understand the paper, and then we define the four metrics used for the performance assessment in the computational results. 

\subsection{Definitions on multiobjective optimization}
A MOCO problem can be defined as
\begin{align} 
 & \min f(x) =  (f_{1}(x),\ldots,f_{p}(x)),  \label{model:mocobegin}\\
 & \textit{ s.t.} \quad  x\in X \subset \mathbb{R}^n, \label{model:mocoend}  %\notag,
\end{align}
where $x$ is the decision vector, $p\in\mathbb{N}$ is the number of objectives with $p\geq2$, $f_{i}:X\rightarrow\mathbb{R}$ with $i=1,\ldots,p$ are the objective functions, and $X\neq \emptyset$ denotes the feasible solution set, which is discrete and bounded.

The notion of optimality with several objective functions is considered in the sense of Pareto optimization. Given two vectors $x,y \in \mathbb{R}^{p}$, 
we say that $x$ is \emph{weakly dominated} by $y$ if $y_i \leq x_i$ for all $i=1,\ldots,p$  (denoted $y \leqq x$). If $y$ weakly dominates $x$ and $y_k< x_k$ for at least one $k\in\left\{ 1,\ldots, p\right\}$, then we say that $y$ \emph{dominates} $x$ (denoted $y~\leq~x$) or $x$ \emph{is dominated} by $y$. 
If strict inequality holds for all $k\in\left\{ 1,\ldots,p\right\} $, then we say that $y$ \emph{strictly dominates} $x$ (denoted $y < x$) or $x$ is \emph{strictly dominated} by $y$. 
 
A feasible solution $x\in X$ is an \emph{efficient solution} if there is no $y\in X$ such that $f(y)$ dominates $f(x)$.
A solution $x\in X$ is called \emph{weakly efficient} if there is no $y\in X$ such that $f(y)$ strictly dominates $f(x)$.
The image of an efficient solution $x$ is a \emph{non-dominated point}, $z=f(x)$. The image of a \emph{weakly efficient solution} $x\textsc{\char13}$ is a \emph{weakly non-dominated point}, $z\textsc{\char13}=f(x\textsc{\char13})$ .
The set of all efficient solutions in a MOCO problem is called \emph{efficient set}, $X_{E}$,  and its image is the Pareto front, \emph{PF}$~=~f(X_{E}) \subseteq \mathbb{R}^p$. 
An efficient solution is \emph{supported} if its image lies on the frontier of the convex hull  of \emph{PF} $\subseteq \mathbb{\mathbb{R}}{}^{p}$. Equivalently, $x\in X$ is supported if it minimizes a weighted sum of the $p$ objectives involving positive weights.
Due to the fact that, in MOCO, many of the elements in $X_{E}$ could lead to the same image, we are only interested in the set \emph{PF} and one anti-image for each element of this set.

Let $z^{1},z^{2} \in \mathbb{R}^{p}$. We say that $z^{1} <_{lex} z^{2}$ if there exists an index $q$, where $z^{1}_{q} < z^{2}_{q}$ and $q = min\{k \mid z^{1}_{k} \neq z^{2}_{k} \}$. The symbol $<_{lex}$ represents the strict \emph{lexicographic order}.

Given a permutation $\sigma$ and a vector function $f:X \rightarrow \mathbb{R}^p$, we denote by  $f_{\sigma}=( f_{\sigma(1)}, f_{\sigma(2)},\ldots,f_{\sigma(p)})$ the vector function where the objectives are reordered using $\sigma$. We say that $x \in X$ is a \emph{lexicographic optimal solution} for permutation $\sigma$ if there is no $y \in X$ with $f_{\sigma}(y) <_{lex} f_{\sigma}(x)$. There exists a maximum of $p!$ different lexicographic optimal solutions, one for each permutation.

Given a Pareto front, \emph{PF}, the \emph{ideal point} is $z^{I}=( z^{I}_{1}, z^{I}_{2}, \ldots, z^{I}_{p})$, where $z^{I}_{i}=$ $\displaystyle \min_{x \in X} f_{i}(x)=$ $\displaystyle \min_{z \in PF} z_{i},$ \,\, $\forall i = 1,\ldots,p $, and the \emph{nadir point} is $z^{N}=( z^{N}_{1}, z^{N}_{2}, \ldots, z^{N}_{p})$, where $z^{N}_{i}=\displaystyle \max_{x \in X_{E}} f_{i}(x)= \max_{z \in PF} z_{i}, \,\, \forall i = 1,\ldots,p $.
It is clear that $z^{I}$ and $z^{N}$ are a lower and an upper bound for the Pareto front. While the ideal point is found by solving $p$ single objective optimization problems, the computation of the nadir point without knowing the whole Pareto front involves optimization over the efficient set, a very difficult problem in general~\cite{ehrgott2003computation}.

Although it is common to use the term `efficient solution' in the decision space and `non-dominated point' in the objective space, sometimes we use the term `solution' to refer to both spaces, when there is no possible confusion.
We assume that $|PF|$ $> 1$, otherwise the problem can be solved in one iteration with a single-objective optimization technique.

Given two vectors $l, u \in \mathbb{R}^p$ with $l<u$, we define the \emph{box} $[l,u]$ as:
\begin{equation} \label{def:box}
[l,u] = \{ x \in \mathbb{R}^{p} \mid l_{i} \leq x < u_{i},  \, \forall i =1,\ldots,p\}.
\end{equation}
When the lower bound of a box is the ideal point in a Pareto front, we usually represent the box only with its upper bound, that is, $[z^{I},u]=[u]$.

\subsection{Performance assessment} \label{sec:performance_assessment}
The performance assessment of algorithms for multiobjective optimization is not a trivial issue. A good analysis comparing different multiobjective evolutionary algorithms is given in~\cite{zitzler2003performance}. Recently, Li and Yao~\cite{li2019quality} published a survey of the quality evaluation for 100 different metrics. In~\cite{jiang2014consistencies}, Jiang et al. categorize the MOO metrics into four groups: capacity metrics, convergence metrics, diversity metrics, and convergence-diversity metrics. Despite the aforementioned difficulty of designing anytime algorithms with good spread (effective anytime algorithms) because performance is often evaluated subjectively (see~\cite{lopez2014automatically}), we have tried to take a representative group of metrics in which we can assume that the better the value of these metrics, the better the performance of the algorithm. Thus, we take one metric from each of the four categories defined by Jiang et al.

Capacity metrics~\cite{jiang2014consistencies} quantify the number or ratio of non-dominated solutions. We use here the Overall Non-dominated Vector Generation Ratio, which is defined as 
\begin{equation} \label{eq:onvgr}
    \text{ONVGR}(N) = \frac{|N|}{|PF|},
\end{equation}
where $N \subseteq PF$ is the set of non-dominated solutions found by a run of a search algorithm and $|\cdot|$ is the cardinality of a set. Thus, ONVGR is a rational number in the interval $[0,1]$ that represents the fraction of points of the Pareto front that were found by the search algorithm. For computing ONVGR, we need to know the total number of points in the Pareto front.

Convergence metrics measure the degree of proximity of the set of solutions to the complete front.
The \emph{additive epsilon indicator}  gives the minimum additive factor by which the approximation set has to be translated in the objective space in order to weakly dominate the reference set~\cite{zitzler2003performance, liefooghe2016correlation}. We have scaled each objective to obtain a value in the range [0,1]. This additive epsilon indicator is defined as  
\begin{equation} \label{eq:epsadd}
    \varepsilon_{+}(N) = \max_{x \in PF} \min_{y \in N} \max_{i = 1,\ldots, p} \left( \frac{y_{i} - x_{i}} {r_{i}} \right),
\end{equation}
where $r_{i}$ is the range of objective $i$ in \emph{N}. This metric is also used in the computational experiments conducted by Ceyhan et al.~\cite{ceyhan2019finding} with the name of \emph{coverage gap}.

\emph{Diversity metrics} indicate the distribution and spread of solutions. From this group, we have selected the \emph{general spread} metric, also cited in~\cite{jiang2014consistencies}. The original \emph{spread} metric (Deb et al.~\cite{deb2000fast}) calculates the distance between two consecutive solutions, which only  works for 2-objective problems. An extension to any dimension, defined in~\cite{zhou2006combining}, computes the distance from a point to its nearest neighbor.
Let $N$ be a set of non-dominated points and  $\{e_{i}\}_{i=1}^{m}$ the extreme solutions\footnote{The original paper does not clarify what they mean by extreme points.}, which are the images of the lexicographic optimal solutions
of the complete Pareto front, $\bS{d}=\frac{1}{\lvert N \lvert} \sum_{x \in N} d(x, N)$, where  $\displaystyle d(x,N)=\min_{y \in N; y \neq x} d^{*}(x,y)$ and $d^{*}$ denotes the Euclidean distance between two $p$-dimensional vectors. We define the \emph{general spread} as
\begin{equation} \label{eq:spread}
    \Delta^{*}(N) = \frac{\sum_{i=1}^{m} d(e_{i},N) + \sum_{x \in N} \lvert d(x,N)- \bS{d} \lvert}{\sum_{i=1}^{m} d(e_{i},N) + \lvert N \lvert \bS{d} }.
\end{equation}

Note that the metric $\Delta^{*}$ is a non-negative number. The lower its value, the more well-spread is $N$. 
Every lexicographic optimal solution not obtained in the execution produces a positive value in $\sum_{i=1}^{m} d(e_{i},N)$ and a higher value for $\Delta^{*}$.

From the group of~\emph{convergence-diversity metrics} we have selected the \emph{hypervolume} indicator.
Given a set of $d$ non-dominated points, \emph{N}$=\{z^{1},z^{2},\ldots,z^{d}\}$, the \emph{hypervolume} $\text{HV}$ is the measure of the region which is simultaneously dominated by \emph{N} and bounded by a reference point $r \in \mathbb{R}^{p}$: 
\begin{equation} \label{eq:hypervolume}
    \text{HV}(N,r)=volume \left( \bigcup_{j=1}^{d} [z^{j},r] \right),
\end{equation}
where $volume$ is the Lebesgue measure in $\mathbb{R}^p$. This is the quality measure with the highest discriminatory power among the known unary quality measures~\cite{lopez2014automatically,rostami2017fast,zitzler2003performance}.
There are many software packages that calculate the exact hypervolume, given a non-dominated set and a reference point~\cite{fonseca2006improved,while2012fast}. The reference point can be taken as  $r_{i} = \displaystyle \max_{j=1,\ldots ,d} z^{j}_{i}$ $\quad \forall i = 1, \ldots , p$. This is an acceptable choice when we have no information about the complete Pareto front.
In the computational experiments presented in this paper, we have calculated the complete fronts for all the instances in order to have a fitted reference point. This means that we know the nadir point for every instance. As noted in the work of Li and Yao~\cite{li2019quality}, there is still no consensus on how to choose a proper reference point for a given problem. To assign some importance to the extreme points, we add one unit to the nadir point, so the reference point is $r_{i} = z^{N}_{i}+1 \quad \forall i=1,\ldots, p$. 
Sometimes, it is useful to consider the percentage of the total hypervolume reached:
\begin{equation} \label{eq:hvr}
\text{HVR}(N,r) = \frac{\text{HV}(N,r)}{\text{HV}(\text{\emph{PF}},r)}.
\end{equation}

Given two subsets of non-dominated points, $A$ and $B$, we write $A \succeq B$ if every $z^{2} \in B$ is weakly dominated by at least one $z^{1} \in A$.
A metric is \emph{Pareto compliant} if, for every two subsets  $A$ and $B$ with $A \succeq B$ and $B \nsucceq A$, the metric value for  $A$ is not worse than the metric value for $B$.
It is desirable for a metric to be Pareto compliant~\cite{zitzler2007hypervolume}.
In other words, it must not contradict the order induced by the Pareto dominance relation. Of the four metrics considered, ONVGR, HV and $\varepsilon_{+}$ are Pareto compliant. This is equivalent to saying that, if we add a new element to the subset of solutions, then ONVGR and HV  do not decrease and  $\varepsilon_{+}$ does not increase. Nevertheless, $\Delta^{*}$ is not Pareto compliant, and we have to be more careful with its analysis.

\section{General formulation for solving MOCO problems} \label{sec:general-formulation}
In this section we describe the framework of Dächert and Klamroth~\cite{dachert2015linear} to solve MOCO problems. The framework can be found in Algorithm~\ref{alg:general_MOCO}. Our proposed algorithm, described in Section~\ref{sec:new_algorithm}, is based on this framework, but we add three modifications that let us get a better spread of solutions over the objective space. The method used by Holzmann and Smith~\cite{holzmann2018solving} is also based on this framework. 
The idea of the method is to maintain a set of search zones, $\mathfrak{U}$, which are $p$-dimensional boxes. Every search zone is defined by its upper bound, because we can consider that the ideal point is the lower bound. 

\begin{algorithm} [ht]
\caption{\emph{General method for MOCO problems}} 	\label{alg:general_MOCO}
\begin{algorithmic} [1]
\REQUIRE $P$
\ENSURE $N$
\STATE $N = \emptyset$  \label{alg:general:line_emptyset}
\STATE $\mathfrak{U} \leftarrow \{U\}$ \label{alg:general:line_initial_box}
\WHILE{($\mathfrak{U} \neq \emptyset$)}
\STATE Select $B \in \mathfrak{U}$ \label{alg:general:line_value_lists}
\IF{ (\emph{P($B$) is feasible})} \label{alg:general:is_feasible}
\STATE $x^{*}\leftarrow$ Optimal solution of $P(B)$
\STATE \emph{N = N $\cup$} $\left\{ f(x^{*})\right\} $
\STATE Update $\mathfrak{U}$ \label{alg:general:line_partition_filtering}
\ELSE
\STATE $\mathfrak{U} \leftarrow \mathfrak{U} - \{B\}$
\ENDIF
\ENDWHILE 
\end{algorithmic}  
\end{algorithm}

At the beginning, the set $N$, containing the non-dominated points, is empty (Line~ \ref{alg:general:line_emptyset}), and the set of search zones contains the initial element $U$ (Line~\ref{alg:general:line_initial_box}), defined by an upper bound of the nadir point. The mathematical program used by the algorithm, $P$, is an input parameter. 
The algorithm then enters a loop while there is at least one zone to analyze. In each iteration of the loop it selects one zone (Line~\ref{alg:general:line_value_lists}), solves the optimization problem, and, if a new non-dominated point is found, it is saved in $N$. 
Every time it finds a new non-dominated point, it updates the set $\mathfrak{U}$ accordingly (Line~\ref{alg:general:line_partition_filtering}), so as to prevent repeated solutions in the future. Another goal of the updating procedure is to reduce the number of search regions at each iteration. More specifically, at least box $B$ is extracted from $\mathfrak{U}$. The algorithm ends because, in MOCO problems, the number of non-dominated points  is finite.

In what follows, we study the relevant elements of Algorithm~\ref{alg:general_MOCO} and analyze different options which may produce a well-spread set of solutions if we treat it as an anytime algorithm.
We focus on three aspects of the general method that we think are important to consider.

\subsection{Mathematical program} \label{sec:setting_model}
The selected mathematical program is one of the key aspects of the algorithm. Not all methods that solve multiobjective problems have the same performance. In fact, those which are based on the $\varepsilon$-constraint method (Ch.~4 of~\cite{ehrgott2005multicriteria}) should not be good as effective anytime algorithms, unless we use a strategy for the selection of the next search zone to explore that guarantees a good dispersion of solutions in the objective space. 
The mathematical program used in the SBA method by Ceyhan et al.~\cite{ceyhan2019finding} is\footnote{We changed the program to adapt it to minimization problems.}:
\begin{align} \label{model:tchebycheff1} 
  &\min_{x \in X} \left ( \varepsilon \sum_{i=1}^{p} w_{i} f_{i}(x) - \alpha \right ),   \\
  & \textit{ s.t.}  \quad \, f_{i}(x) \leq u_{i} - \alpha, \quad i =1,\ldots,p,   \\
  & \qquad \quad \alpha \geq 0, \label{model:tchebycheff2}
\end{align}
where $\alpha$ is a variable which represents the coverage gap,  
$\varepsilon$ is a sufficiently small positive constant and $\omega_{i} > 0 \,\, \forall i$. If the program~\eqref{model:tchebycheff1} to \eqref{model:tchebycheff2} has a solution, then it is efficient. 

Another mathematical program, used in the work of Holzmann and Smith~\cite{holzmann2018solving}, is based on a weighted augmented Tchebycheff norm:
\begin{align} \label{model:holzmann} 
 & \min_{x \in X} \left (
   \max_{ i = 1,\ldots,p} \left( \omega_{i}  \left \lvert f_{i}(x) - z_{i}^{I} \right \lvert  \right)
  + \epsilon \cdot \sum_{i=1}^{p} \omega_{i}  \left \lvert f_{i}(x) - z_{i}^{I}  \right \lvert  \right ) , 
\end{align}
where $z^{I}$ is the ideal point of the problem. The parameters $\omega$ and $\epsilon$ are chosen to make the program  feasible. Depending on the objective value, we conclude whether the problem has a new solution inside the box or not. The reader is referred to the original paper~\cite{holzmann2018solving} for more information.

We observed relevant performance differences between the two programs in a preliminary study. The weighted augmented Tchebycheff norm behaves better regarding the spread of the solutions in the objective space.  

\subsection{Selection of the next search zone to explore} \label{sec:next_search_explore}
Line \ref{alg:general:line_value_lists} of Algorithm \ref{alg:general_MOCO} proposes the  extraction of the next search zone to analyze. The selection may be done randomly, but intuitively, if we split the objective space into different regions, and at each step we explore those which have a higher volume, we expect to obtain a final better spread of the solutions. Following this idea, we can define a numerical value for each search zone, called \emph{priority}. One example of the \emph{priority} function is the volume of the $p$-dimensional hyperrectangle with opposite vertices being the vector $u$ (upper bound for the search zone) and $z^{I}$ (ideal point). For instance, if $u=(10,20,30)$ and $z^{I}=(5,12,20)$, then the volume of the box is $(10-5) \cdot (20-12) \cdot (30-20)=400$, and this could be its priority.
A more sophisticated way of selecting the next box to explore is mentioned in Section~\ref{sec:alternation}.

\subsection{Updating the set of search zones} \label{sec:updating_search_zones_full}
This may be the most important aspect to take into account in Algorithm~\ref{alg:general_MOCO} (Line \ref{alg:general:line_partition_filtering}) in terms of  termination of the general algorithm and computational effort.
One way to ensure that a non-dominated point is never computed again during the search consists in splitting the objective search zone and discarding at least that point. A good way to do this is found in the work of Dächert and Klamroth~\cite{dachert2015linear}. 
After a solution $z$ is found, they divide the search zone into $p$ new zones. The union of the new search zones discards the points dominated by $z$ and the new search regions to explore contain at least one point less than the original search region.
Given a box $B$ with upper bound $u=(u_{1},\ldots,u_{p})$, let us suppose that we solve the mathematical problem and obtain a new non-dominated point $z=(z_{1},\ldots,z_{p})$. Then, we split the box $B$ into $\left\{ B_{i} \right \}_{i=1}^{p}$ with upper bounds $\left\{ u^{i} \right \}_{i=1}^{p}$, being $u^{i}=\left(u_{1},\ldots,u_{i-1},z_{i} ,u_{i+1},\ldots,u_{p} \right)\forall i=1,\ldots,p$. 
Each box $B_i$ with upper bound $u^{i}$ is said to have \emph{direction} $i$. If the lower bound of all the boxes $B$ and $B_i$ is the ideal point, $z^I$, we call the split \emph{full p-split}.
A graphical example is shown in Figures~\ref{fig:boxes_a} and~\ref{fig:boxes_b} for $p=3$.  
Note that when a new non-dominated point is obtained, we must apply \emph{full p-split} to all search zones that contain that new point, in order to assure the non-duplicity of the solutions. \emph{Full p-split} is used in the work of Holzmann and Smith~\cite{holzmann2018solving}. In Section~\ref{sec:updating_search_zone_p-partition} we describe a new way of splitting the objective space, called \emph{p-partition}.

The splitting process often generates redundant zones. If we have two search zones $u^{1},\, u^{2}$ with $u^{1} \leq u^{2}$, all potential non-dominated points generated by $u^{1}$ could also be generated by $u^{2}$, which means that $u^{1}$ is redundant. Therefore, a filtering process should be implemented after the split. Klamroth et al.~\cite{klamroth2015representation} proposed two different algorithms for this purpose. The first is called \emph{RE} (\emph{redundancy elimination}) and consists in eliminating those redundant search zones at each iteration. 
The second algorithm is called \emph{RA} (\emph{redundancy avoidance}) and is based on structural properties of local upper bounds which yield necessary and sufficient conditions for a candidate local upper bound to be non-redundant. The filtering step is avoided in this case. In the computational experiments of the work by Klamroth et al.~\cite{klamroth2015representation}, they obtained better results in runtimes using \emph{RE} when $p \in \{3, 4, 5\}$, and using \emph{RA} when $p>5$. 

A more recent paper by Dächert et al.~\cite{dachert2017efficient} describes a specific neighborhood structure among local upper bounds. With this structure, updates to the search region when a new non-dominated point is found can be more efficient compared to \emph{RE} and \emph{RA} approaches, as the number of points increases, but \emph{RA} and even \emph{RE} perform better for a small number of solutions. 
Since our proposal is designed as an anytime algorithm where the decision maker can stop the execution whenever desired, it is reasonable to think that the total number of solutions may not be high, unless a complete execution is derived. Thus, we have decided to use the \emph{RE} approach in this paper. 

\section{Proposed anytime algorithm: {\AMOCO}} \label{sec:new_algorithm}
In this section, we propose a new exact anytime algorithm based on the general framework described in Section~\ref{sec:general-formulation} but with three novel contributions.
This algorithm can solve any MOCO problem and it can also be used for MODO problems with finite feasible sets. We call it {\AMOCO} because it uses a Tchebycheff mathematical program with Partition of the objective space using Alternating of directions.
The pseudocode is shown in Algorithm~\ref{alg:AMOCO}.

\begin{algorithm} [!ht]
\caption{\AMOCO} 
\label{alg:AMOCO}
\begin{algorithmic} [1]
\REQUIRE $f$ and $X$ \quad // The MOCO problem
\ENSURE $XZ$ \quad \quad // A set of efficient solutions and their image
\STATE $\mathfrak{L} = \left( \mathfrak{L}_{1}, \ldots, \mathfrak{L}_{p}\right )$ \quad and \quad  $\mathfrak{L}_{i} = \emptyset , \quad \forall i=1, \ldots, p$ \label{alg:AMOCO:init_L}
\STATE $XZ = \emptyset$ \label{alg:AMOCO:init_PF}
\STATE $k=0$\label{alg:amoco:init_counter}
\STATE Compute the ideal point and an estimation of the nadir point: $z^{I}$ and $\overline{z}^{N}$ \label{alg:AMOCO:get_initial_bounds}
\STATE Set $\epsilon = 1 / (2p(r-1))$ where $r = \max_{i=1}^{p} \left \{ \overline{z}^{N} - z^{I} \right \}$ \label{alg:amoco:set_epsilon}
\STATE $\mathfrak{L}_{1}.insert \left ( \left [z^{I},\overline{z}^{N} \right ]  \right )$ \label{alg:amoco:l1_insert}
\STATE $P(u,\omega) \equiv \displaystyle \min_{x \in X} \left (
   \max_{ i = 1, \ldots,p} \left( \omega_{i}   \left \lvert f_{i}(x) - z_{i}^{I}  \right \lvert   \right)
  + \epsilon \cdot \sum_{i=1}^{p} \omega_{i}  \left \lvert f_{i}(x) - z_{i}^{I} \right \lvert \right )  $ 
\label{alg:amoco:P}
\WHILE{$\exists \, i: |\mathfrak{L}_{i}| > 0$ \AND  $intime$}
\STATE $(B,k) \leftarrow$ Select\_next\_box($\mathfrak{L}$, $k$) \label{alg:select_next_box} 
\STATE Set $\omega = (\omega_{1},\ldots,\omega_{p})$ where $\omega_{i} = 1 / \max \{1 , B.u_{i} - z^{I}_{i} \}$ \label{alg:amoco:set_w}
\STATE $(x^*,obj)$ $\leftarrow$ Solve $P(B.u,\omega)$
\label{alg:amoco:solve_model} 
\IF{ $obj < 1$ \AND  $z^{I} < B.u$}
\label{alg:amoco:feasible}
\STATE \emph{XZ $\leftarrow$ XZ $\cup$} $\left\{ \left(x^{*},f(x^{*})\right) \right\}$  \label{alg:amoco:add_solution}
\STATE \emph{Update} $ \left (\mathfrak{L},f(x^{*}) \right )$ \label{alg:amoco:update}
\ELSE
\STATE $\mathfrak{L}_{k}.remove(B)$ \label{alg:amoco:remove_box}
\ENDIF
\ENDWHILE 
\end{algorithmic}  
\end{algorithm}

We now provide a high-level description of the algorithm and go in depth into the novel contributions in separate subsections. Every search zone is a box with the following fields: \emph{lower bound} ($l$), \emph{upper bound} ($u$), and \emph{priority}.
The vector $\mathfrak{L}$ contains $p$ priority queues (heaps) of boxes. The boxes in queue $\mathfrak{L}_{i}$ have direction $i$ and are sorted by non-increasing priority. The initial box has as lower and upper bounds the ideal point and an upper bound of the nadir point. Thus, it contains the whole Pareto front. We insert this box into $\mathfrak{L}_{1}$ (Line~\ref{alg:amoco:l1_insert}), although any other queue $\mathfrak{L_{i}}$ could be chosen. The $\epsilon$ value is constant during the execution and depends on the initial bounds (Line~\ref{alg:amoco:set_epsilon}).

{\AMOCO} then executes a loop while there is a box in any list of $\mathfrak{L}$ and the time limit is not reached (Boolean \emph{intime}). In the loop, it first selects the next box to analyze (Line~\ref{alg:select_next_box}) using Algorithm~\ref{alg:next_box}. 
Then, it solves the mathematical program in Eq.~\eqref{model:holzmann} (Line~\ref{alg:amoco:P}). 
As pointed in Section~\ref{sec:general-formulation}, this mathematical program provides better spread of the non-dominated points. The parameters are taken from the work of Holzmann and Smith~\cite{holzmann2018solving}: $\omega_{i} = 1  / \max \{1,u_{i} - z_{i}^{I}  \} , \, \forall i$, and $\epsilon = 1 / (2p(r -1))$, where $u$ is the upper bound of the next box to analyze,
$r = \max_{i = 1,\ldots,p} (\overline{z}^{N}_{i}-z^{I}_{i}) $, $z^{I}$ is the ideal point and $\overline{z}^{N}$ is an upper bound of the nadir point, computed as follows:
\begin{equation} \label{eq:bound_nadir}
    \overline{z}^N_{i} = \max_{x \in X}(f_{i}(x)) \qquad \forall i = 1,\ldots,p.
\end{equation} 
When $u \in \left [ \overline{z}^{N} \right]$, the model is always feasible. Moreover, if $z^{I} < u$ the solution is inside the box $[u]$ if and only if the objective value is less than one unit (Theorem 5 of~\cite{holzmann2018solving}). Thus, we can slightly modify Line~\ref{alg:general:is_feasible} of Algorithm~\ref{alg:general_MOCO}, changing the expression `\emph{is feasible}' by `$obj < 1$ and $z^{I}< u$' (Line~\ref{alg:amoco:feasible} of Algorithm~\ref{alg:AMOCO}).
If it finds a new solution, $x^{*}$, it adds the pair $(x^{*},f(x^{*}))$ to $XZ$ (Line~\ref{alg:amoco:add_solution}) and updates the boxes in the priority queues $\mathfrak{L}$ (Line~\ref{alg:amoco:update}). Otherwise, it discards the box from the corresponding queue (Line~\ref{alg:amoco:remove_box}).

The correctness and termination of {\AMOCO} is proven in 
Theorem~\ref{theorem:tpa_works}.
This is an exact anytime algorithm and, thus, the complete Pareto front is found if the algorithm is given enough time. This guarantees the optimality and convergence of the algorithm.
Observe that all the solutions found belong to the Pareto front. This contrasts with other heuristic and metaheuristic algorithms that can only compute approximate solutions without any guarantee of being efficient. Such algorithms require mechanisms to explore the search space that can be interpreted differently by different developers. Rostami et al.~\cite{rostami2020algorithmic} have recently studied this issue and found that the resulting approximate solutions can influence the final results in such a way that the differences in quality metrics can even be statistically significant.
The reliability and stability of {\AMOCO} are supported by the ILP solver used. The memory usage and computation time are proportional to the number of boxes to maintain, and this can be exponential in the number of solutions found. Furthermore, the number of solutions in the Pareto front can be exponential in the number of decision variables. Thus, the memory required by the algorithm and the run time can be exponential in the number of variables if the whole Pareto front is computed. Regarding memory usage, in our experiments, using 2 GB of RAM was enough even when the complete Pareto front was computed. 

In the next subsections we will detail the three novel contributions of the algorithm: a method to diversify the search exploring boxes with different directions, a new way to split the boxes forming a partition of the original one, and a new priority function for the boxes to explore. 

\subsection{Alternation of directions in the search} \label{sec:alternation}

In this section we propose a new strategy for the selection of the next box to explore, trying to diversify the regions of the objective space that are explored. This is the first novel contribution of this paper.
The priority queue $\mathfrak{L}_{i}$ contains boxes with \emph{direction}~$i$. Boxes in different directions are displaced in the objective space along the different dimensions. We propose to select a box from a different priory queue $\mathfrak{L}_{i}$ in each iteration of {\AMOCO}. Among the ones with the same direction {\AMOCO} chooses the box with higher \emph{priority}. The pseudo-code of the Select\_next\_box procedure (Line~\ref{alg:select_next_box} in Algorithm~\ref{alg:AMOCO}) is in Algorithm \ref{alg:next_box}.

\begin{algorithm} [!ht]
\caption{Select\_next\_box} 	
\label{alg:next_box}
\begin{algorithmic}[1]
\REQUIRE $\left \{ \mathfrak{L}_{i} \right \}$, $k$ \qquad  // List of boxes and integer counter
\ENSURE $(B, k)$
\STATE $B=\emptyset$ 
\IF{$\exists \, i: |\mathfrak{L}_{i}| > 0$}
\STATE $k=(k  \mod  p) + 1$
\WHILE{$|\mathfrak{L}_{k}| = 0 $}
\STATE $k=(k  \mod  p) + 1$  
\ENDWHILE
\STATE $B \leftarrow \text{argmax}_{B \in \mathfrak{L}_{k}} \left(B.priority \right )$
\ENDIF
\end{algorithmic}
\end{algorithm}

\subsection{New updating procedure of the search zone} \label{sec:updating_search_zone_p-partition}

This section presents a new way of splitting the box, called \emph{p-partition}. This is the second contribution of the paper. The main difference between \emph{p-partition} and \emph{full p-split} is that the new boxes created by \emph{p-partition} form a partition of the original box (they are pairwise disjoint).

\begin{definition} \label{def:partition}
Let $B=[l,u]$ be a box and $z \in \mathbb{R}^p$ a point. 
For every $i=1,\ldots,p$, we define $B_{i}(z)=\left \{x \in B \mid x_{i} < z_{i}  \text{ and }  x_{j} \geq z_{j} \, \forall j >  i  \right \}$.  We also define $B_{0}(z)$=$\{x \in B $ $\mid x_{j}$ $\geq z_{j}$ $\, \forall j  \}$. 
\end{definition}

It is easy to check that  $B_i(z)$ are boxes, which can be written as follows:
 \begin{equation} \label{eq:boxbk_}
        B_{i}(z) = \left[
        \left(l_{1},\ldots,l_{i},\hat{z}_{i+1},\ldots,\hat{z}_{p}\right) ,
        \left( u_{1},\ldots,u_{i-1},\hat{z}_{i},u_{i+1},\ldots,u_{p} \right)
        \right],
\end{equation}
where $\hat{z}_{i} = \max \{z_{i}, l_{i} \}$ $\forall i=1,\ldots,p$. 
If $B_i(z) \neq \emptyset$, then it has direction $i$.

\proposition{Given a box $B$, then $\left\{B_{i}(z) \right \}_{i=0}^{p}$ is a partition of $B$.} \label{proposition:partition}
\begin{proof}
We need to prove that $B_{i}(z) \cap B_{j}(z) = \emptyset \,\,  \text{for all } i \neq j$ and  $\bigcup\limits^p_{i=0} B_{i}(z) =B$.
Suppose $x \in B_{i}(z) \cap B_{j}(z)$ for indexes $i,j$ with $i<j$. As $x \in B_{i}(z)$, then $x_{j} \geq z_{j}$. As $x \in B_{j}(z)$, then $x_{j} < z_{j}$. This is a contradiction, so the sets are always disjoint and the first part is proved. 

Let $x \in \bigcup\limits^p_{i=0} B_{i}(z)$. By definition, $x \in B$, so $\bigcup\limits^p_{i=0} B_{i}(z) \subset B$. Now let $x \in B$. If $x_{i} \geq z_{i} \,\, \forall i$, then $x \in B_{0}(z)$. Otherwise, let $j$ be the higher index which verifies $x_{j} < z_{j}$. Then, $x \in B_{j}(z) \subset \bigcup\limits^p_{i=0} B_{i}(z)$, so 
$\bigcup\limits^p_{i=0} B_{i}(z) =B$ and the second part is proved. 
\end{proof}

\begin{definition} \label{def:p-partition}
Let $B=[l,u]$ be a box, and $z < B.u$. We define the  \emph{p-partition} of the box $B$ according to $z$ as $\left\{B_{i}(z) \right \}_{i=1}^{p}$.
\end{definition}

When obtaining a non-dominated point $z$, the exploration of some other boxes in the future might produce the same solution. In order to prevent this, we split those boxes and remove the space weakly dominated by $z$, which is $B_0(z)$. That is why \emph{p-partition} does not include $B_0(z)$.

\begin{figure} [!ht]
\begin{minipage} {0.47\linewidth}
\begin{center}
\includegraphics[scale=0.18]{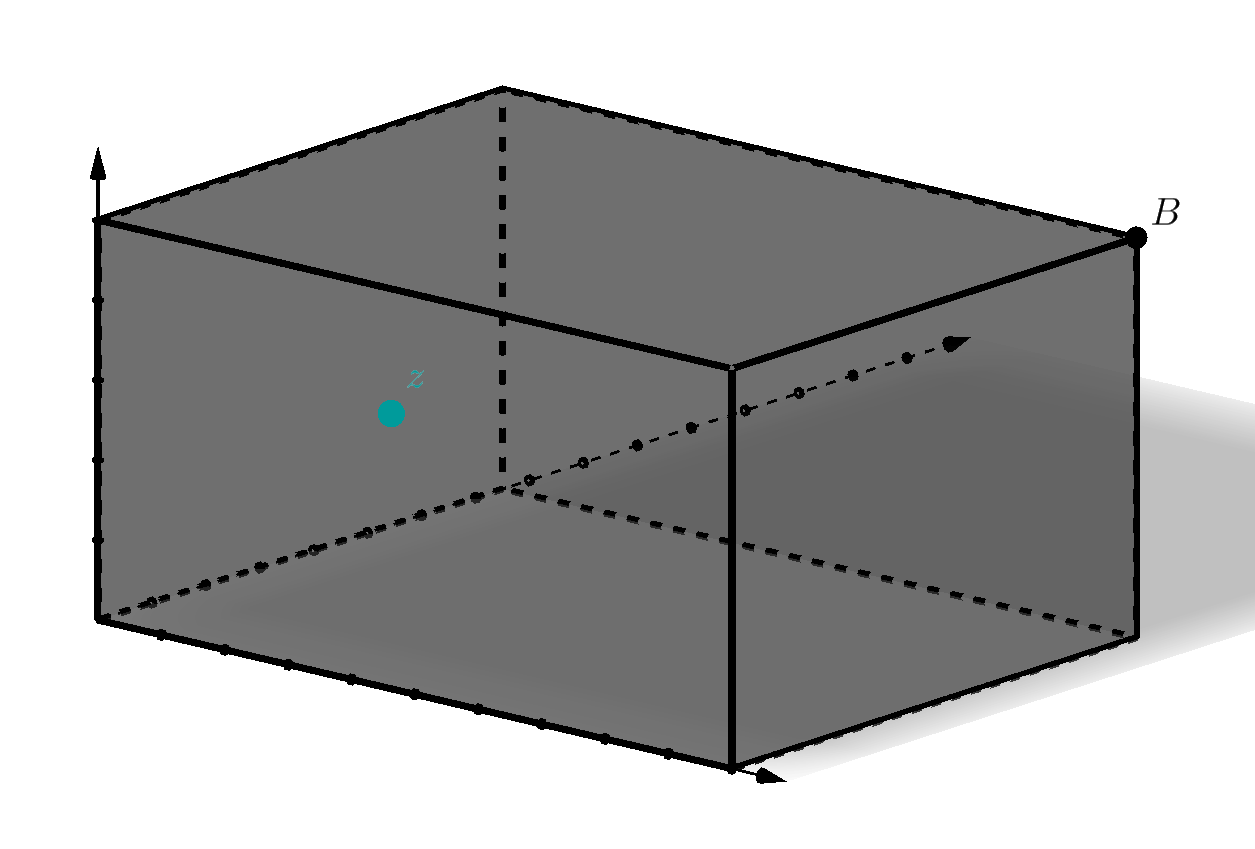}
\end{center}
\subcaption{Box $\mathbf{B}=[(0,0,0),\,(20,15,10)]$ with non-dominated point $z=~(5,5,5)$ inside. \\ \\ \\}
\label{fig:boxes_a}
\end{minipage}
\begin{minipage} {0.47\linewidth}
\begin{center}
\includegraphics[scale=0.18]{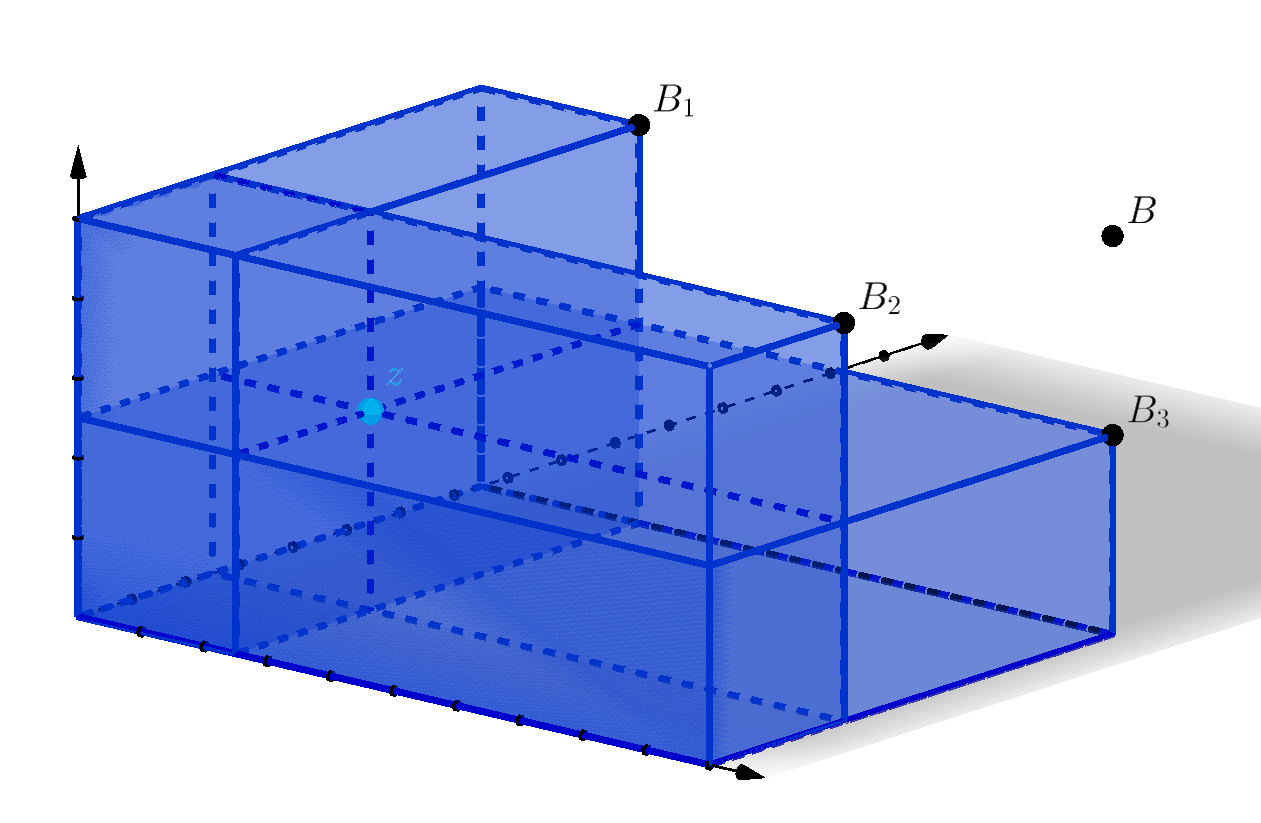}
\end{center}
\subcaption{Resulting boxes after \emph{full p-split}: $\mathbf{B_{1}}=~[(5,15,10)]$, $\mathbf{B_{2}}=~[(20,5,10)]$, and  $\mathbf{B_{3}}=~[(20,15,5)]$. All the lower bounds are $(0,0,0)$. Each box shares some space with another box.}
\label{fig:boxes_b}
\end{minipage}
\begin{minipage} {0.99\linewidth}
\begin{center}
\includegraphics[scale=0.2]{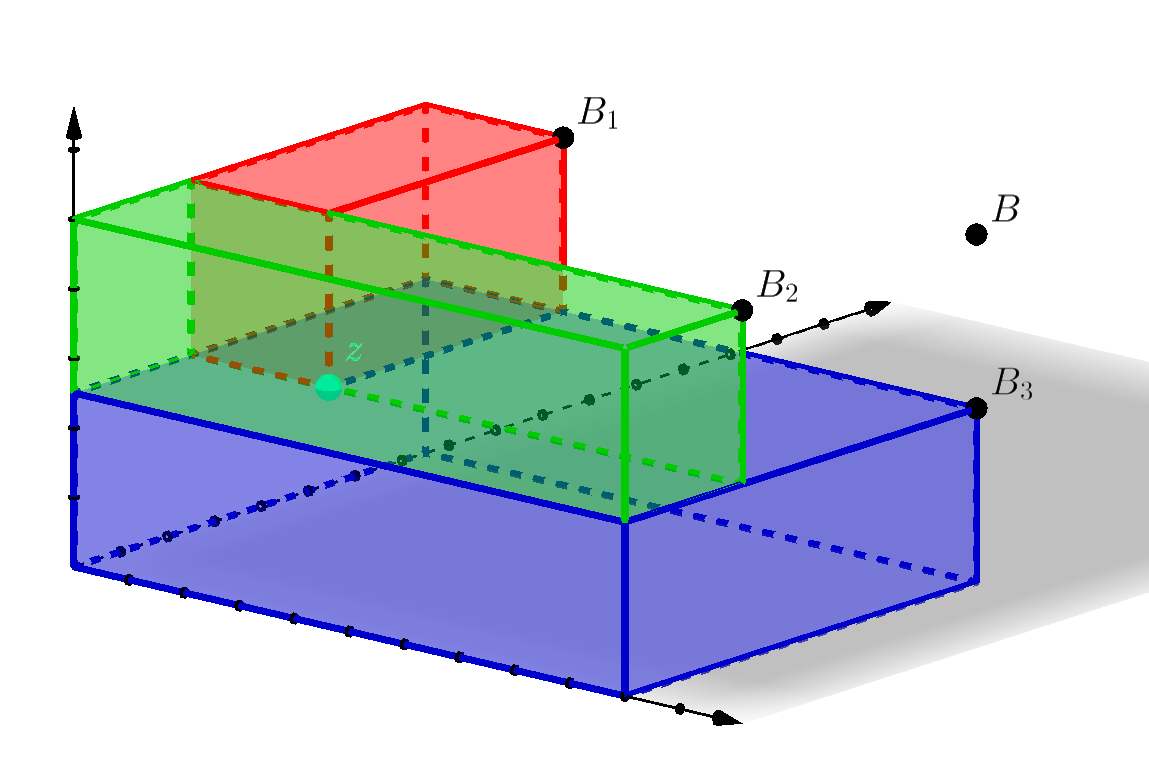}
\end{center}
\subcaption{Resulting boxes after \emph{p-partition}: $\mathbf{B^{*}_{1}}=[(0,5,5),\,(5,15,10)]$,  $\mathbf{B^{*}_{2}}$$=[(0,0,5),\,(20,5,10)]$, and $\mathbf{B^{*}_{3}}=[(0,0,0),\,(20,15,5)]$. Each pair of boxes is disjoint. } 
\label{fig:boxes_c}
\end{minipage}
\caption{Splitting a box using \emph{full p-split} and \emph{p-partition}. Upper bound for the new boxes is the same in both cases, but lower bounds change.}
\label{fig:boxes}
\end{figure}

Figure~\ref{fig:boxes} shows a graphical example of \emph{p-partition} compared to \emph{full p-split}. Note that the upper bounds in the resulting boxes are the same, but lower bounds are different.

It is very important to eliminate redundant boxes because, otherwise, the computational execution time grows exponentially. 
For \emph{p-partition}, we apply a variant of the redundancy elimination (\emph{RE}) used in \emph{full p-split}.
In the \emph{RE} method, when a box $A$ is contained in a box $B$, we eliminate $A$ from the priority queue of boxes to explore. For a deep study of how this method works, see~\cite{klamroth2015representation}. Nevertheless, it is not possible to apply this idea to \emph{p-partition} because the lower bounds of the new boxes differ. We join boxes to save this obstacle.

\begin{definition}
\label{def:join}
Let $A=[l^{a},u^{a}]$ and $B=[l^{b},u^{b}]$ be two boxes with $u^{a} \leq u^{b}$. We define the \emph{join box} of the two boxes as a new box $C=[l^{c},u^{c}]$, where $l^{c}_{i} = \min \{l^{a}_{i},l^{b}_{i}\}$ and $u^{c}_{i} = u^{b}_{i}$ $\forall i=1,\ldots,p$.
\end{definition}

Once we introduce join boxes, we cannot guarantee that all the boxes are pairwise disjoint, but we can confirm that the new box does not exclude non-dominated points. We prove this in the next proposition.
The \emph{priority} of the new box $C$ could be greater than the value of the previous boxes $A$ and $B$. This means that box $C$ will have higher preference to be analyzed in subsequent iterations.

\begin{proposition} \label{proposition:reduced_and_not_exlcude}
With the conditions of Definition~\ref{def:join}, it holds that $A \cup B \subset C$, which means that analyzing box $C$ substitutes the analysis of the other two boxes. 
\begin{proof}
Let $x \in A \cup B$. If $x \in A$, then $l^{c}_{i} \leq l^{a}_{i} \leq x < u^{a}_{i} \leq u^{b}_{i} = u^{c}_{i}$ $\, \forall i$, so $x \in C$. If $x \in B$, then $l^{c}_{i} \leq l^{b}_{i} \leq x < u^{b}_{i} = u^{c}_{i}$ $\, \forall i$, so $x \in C$.
\end{proof}
\end{proposition}

We are now ready to present in Algorithm~\ref{alg:update} the updating procedure of Line~\ref{alg:amoco:update} in Algorithm~\ref{alg:AMOCO}. It is divided into two parts. 
In the first part (Lines~\ref{alg:part2_begin} to~\ref{alg:part2_end}), the boxes in $\mathfrak{L}$ are split using \emph{p-partition}. Some of the new boxes may be empty. The non-empty boxes  are inserted into the corresponding $\mathfrak{L}_{i}$ according to their direction (Line~\ref{alg:update:insert_li}). The priority function used for the boxes is detailed in the next subsection.
The second part (Lines~\ref{alg:part3_begin} to~\ref{alg:part3_end}) filters the redundant boxes using \emph{RE} but taking into account that if a domination relation between two upper bounds is detected, the boxes are joined and the new box replaces the older ones (Lines~\ref{alg:part3_join_begin} to~\ref{alg:part3_join_end}). We finish this subsection proving the correctness and termination of {\AMOCO}.

\begin{proposition} \label{proposition:not_twice}
Algorithm {\AMOCO} never finds the same non-dominated point twice.
\begin{proof}
Let $B$ be the box being explored in {\AMOCO} and $z$ the non-dominated point found solving $P(B.u,\omega)$ in Line~\ref{alg:amoco:solve_model} of Algorithm~\ref{alg:AMOCO}. The definition of the mathematical program $P$ ensures that $z < B.u$, and the use of \emph{p-partition} with each box $B'$ such that $z < B'.u$ discards $B'_0(z)$, which is the only box that can contain $z$. 
The upper bound of a join box is always the upper bound of a previous box. 
Thus, after the updating procedure, the remaining boxes do not contain $z$ and $z$ cannot be found again. 
\end{proof}
\end{proposition}

\begin{algorithm} [!ht]
\caption{Update}
\label{alg:update}
\begin{algorithmic} [1]
\REQUIRE $(\mathfrak{L},z)$
\ENSURE $(\mathfrak{L},z)$
\FORALL {$B \in \bigcup_{k=1}^{p} \mathfrak{L}_{k}$ with $z < B.u$} \label{alg:part2_begin}
\FOR {$i = 1$ \TO $p$}
\STATE $B_{i}(z)=\left \{x \in B \mid x_{i} < z_{i} \text{ and } x_{j} \geq z_{j} \, \forall j >  i  \right \}$
\IF {$\left(B_{i}(z) \neq \emptyset \right)$}
\STATE $B_{i}(z).priority \leftarrow reduced\_scaled(B , i ,z)$
\STATE $\mathfrak{L}_{i}.insert(B_{i}(z))$ \label{alg:update:insert_li}
\ENDIF
\ENDFOR
\ENDFOR  \label{alg:part2_end}
\FOR {$i = 1$ \TO $p$} \label{alg:update:for_loop_filtering} \label{alg:part3_begin}
\FORALL {$ A, B \in \mathfrak{L}_{i} \text{ with } A.u \leq B.u $} 
\STATE $C \leftarrow join(A,B)$ \label{alg:part3_join_begin}
\STATE $\mathfrak{L}_{i}.remove(A)$ 
\STATE $\mathfrak{L}_{i}.remove(B)$ 
\STATE $\mathfrak{L}_{i}.insert(C)$ 
\label{alg:part3_join_end}
\ENDFOR
\ENDFOR  \label{alg:part3_end}
\end{algorithmic}  
\end{algorithm}

\begin{theorem} \label{theorem:tpa_works}
{\AMOCO} terminates after finding all the non-dominated points of the MOCO problem if the time limit is not set.
\begin{proof}
     The number of non-dominated points is finite in MOCO problems. By Proposition~\ref{proposition:not_twice}, no points are found twice. Thus, {\AMOCO} terminates. Proposition~\ref{proposition:reduced_and_not_exlcude} ensures that all \emph{PF} is found if enough time is given.
\end{proof}
\end{theorem}

\subsection{A new priority function}
\label{sec:priority_value}

As the third contribution, we define a new priority function for the boxes. We want to give more priority to boxes having a larger region where non-dominated points can potentially exist. Given a non-empty box $B=[l,u]$, we define the $scaled$ function as 
\begin{equation} \label{eq:scaled_2}
scaled(B)=\prod_{i=1}^{p} \left (\frac{u_{i}-l_{i}}{\overline{z}^{N}_{i}-z^{I}_{i}} \right ).
\end{equation}

Note that $scaled(B) > 0$. In the case of an empty box, we define $scaled(B)=0$. We assume, without loss of generality, that $\overline{z}_{i}^{N} > z_{i}^{I} \,\, \forall i$, otherwise the objective $i$ always takes the same value and can be discarded.

In some boxes, there is a region that cannot contain any non-dominated point. In this case, the $scaled$ function overestimates the priority of the box. We define a new priority function to correct this overestimation. First, we need some results.

\proposition{Let $B=[l,u]$ be a box and $z \in B$, then the box $[l,z] \subseteq B_{p}(z)$, where \emph{p} is the dimension of the objective space}.
 \label{proposition:discarded_zone}
 \begin{proof}
 From the definition of box, $z < u$. 
 Let $x \in [l,z]$. Then, $l_{i} \leq x_{i} < z_{i} < u_{i} \,\, \forall i = 1, \ldots, p$. More specifically, $x_{p} < z_{p}$. This means that $x \in B_p(z)$, so $[l,z] \subseteq B_{p}(z)$.
\end{proof}

If $z = l$, we have $[l,z] = \emptyset$, and $B_{i}(z)=\emptyset$ for all $i=1,\ldots,p$. The box $[l,z]$, if it is not empty, contains points dominating $z$. Thus, if $z$ is non-dominated, the box $[l,z]$ does not contain any non-dominated point, otherwise $z$ is dominated and we have a contradiction.

\begin{definition}
Let $B = [l,u]$ be a box, $z < B.u$ a non-dominated point and $B_{i}(z)$ for $i=1,\ldots,p$ the box with direction $i$ after \emph{p-partition}. Then, for every $i=1,\ldots,p$, we define the \emph{reduced\_scaled} function as
\begin{equation}
    \small
    reduced\_scaled(B, i, z)= \left\{ 
    \begin{array}{lcc}
     scaled(B) &   \text{if}  & i \neq p \\
     scaled(B) - scaled([l,z]) &  \text{if} & i = p.
     \end{array}
   \right.
\end{equation}
\end{definition}

Note that some of the boxes after \emph{p-partition} may be empty.

\proposition{The \emph{reduced\_scaled} value is always non-negative.}
\begin{proof}
For $i=1,\ldots,p-1$, it is obvious that $reduced\_scaled$ is non-negative (see Eq.~\eqref{eq:scaled_2}). Let us now assume that $i=p$. If $[l,z]=\emptyset$, then $scaled([l,z])=0$ and $reduced\_scaled(B_{p}(z),$ $p,z)$ $=$ $scaled(B_{p}(z)) \geq 0$. 
If $[l,z] \neq \emptyset$, then $z \in B$ and Proposition~\ref{proposition:discarded_zone} ensures that $[l,z] \subseteq B_{p}(z)$, so $scaled([l,z]) \leq scaled(B_{p}(z))$, and $reduced\_scaled$ is non-negative.
\end{proof}

We use the $reduced\_scaled$ function as the \emph{priority} value of a box. 

\section{Computational results} \label{sec:computational_results}
In this section, we present a deep experimental study divided in four subsections. The first subsection describes the benchmark of instances we have used. The instances are grouped into categories. Each category has a fixed number of variables. The categories are also grouped into classes, where each class corresponds to a different MOCO problem. The authors have considered the choice of this benchmark appropriate because different MOCO problems could have different performances. 
The second subsection describes the parameters for the three algorithms used in the comparison. 
The third subsection shows the results of the computational experiments. For each instance and algorithm, the results are displayed at some time points.  We provide a deep analysis of the results in the last subsection, including a statistical validation. 

\subsection{Benchmark}
The benchmark we use is divided into three classes of well-known multiobjective problems. The first is the multiobjective knapsack problem (\emph{KP}), the second is the multiobjective assignment problem (\emph{AP}), and the third class corresponds to multiobjective integer linear programming problems (\emph{ILP}). All the details of the formulation of the three classes can be found in~\cite{kirlik2014new}, at URL \url{http://home.ku.edu.tr/~moolibrary/}, and in the supplementary material.
For each class, different problem categories are generated based on problem size. There are 10 instances in each category. In total, there are 480 instances.
As stated in Section \ref{sec:background}, the metrics we use to compare the algorithms are ONVGR, HV, $\Delta^{*}$, and $\varepsilon_{+}$. To calculate these quality indicators, we need to calculate the complete Pareto front. The details of these experiments are shown in the supplementary material of the paper.

\subsection {Algorithms and parameters}
We downloaded and used the source code\begin{footnote} {Commit 6b3b7e7bb9 on 17 Aug 19 at  \url{https://github.com/gokhanceyhan/MOIP\_Solvers}.} \end{footnote} provided by Ceyhan et al.~\cite{ceyhan2019finding}. This algorithm will be called \emph{ceyhan} henceforth. The algorithms of Holzmann and Smith~\cite{holzmann2018solving} (called \emph{holzmann}) and {\AMOCO}\begin{footnote} {Source code available at \url{https://github.com/MiguelAngelDominguezRios/boxMO}. \,} \end{footnote} were programmed using C++.
Since \emph{holzmann} and {\AMOCO}, are taken from the same framework, we use in them the same filtering process of the solutions, the \emph{RE} method, in order to make a fair comparison and observe the differences of performance with our new contributions. 

The computer used for the experiments is a multicore machine with four Intel Xeon CPUs (E5-2670 v3) at 3.1~GHz, a total of 48 cores, 64 GB of memory and Ubuntu 16.04 LTS. For each run we used only 1 core and 2 GB of RAM. 

The three algorithms use CPLEX 12.6.2 as the ILP solver. We set the CPLEX parameters, \begin{footnotesize} CPXPARAMEPGAP \end{footnotesize} $=$ \begin{footnotesize} CPXPARAMEPAGAP \end{footnotesize} $=$ \begin{footnotesize} CPXPARAMEPINT  \end{footnotesize} $=0$ and \begin{footnotesize}  CPXPARAMPARALLELMODE \end{footnotesize} $=$ \begin{footnotesize} CPXPARAMTHREADS  \end{footnotesize} $= 1$. Parameter \begin{footnotesize} CPXPARAMEPINT \end{footnotesize} indicates the integrality tolerance for the solution variables. Its default value is $10^{-5}$. Parameter \begin{footnotesize} CPXPARAMEPGAP \end{footnotesize} sets a relative tolerance on the gap between the best integer objective and the objective of the best node remaining in the tree used in the \emph{branch and cut}. The default value is $10^{-4}$. \begin{footnotesize} CPXPARAMEPAGAP \end{footnotesize} sets an absolute tolerance on the gap between the best integer objective and the objective of the best node remaining. The default value is $10^{-6}$. \begin{footnotesize} CPXPARAMPARALLELMODE \end{footnotesize} is set to deterministic mode, which means that multiple runs with the same model and the same parameter settings on the same platform will reproduce the same solution path and results. Finally, \begin{footnotesize} CPXPARAMTHREADS \end{footnotesize} sets the default number of parallel threads that will be invoked by any CPLEX parallel optimizer. In this case, it is set to 1.

Although the three algorithms are deterministic and the time limit is fixed, the number of solutions found can differ in different executions because CPLEX manages some internal parameters, such as the remaining available memory of the machine, which can influence the tree exploration to obtain the next solution. 
To soften these differences, we did 30 executions for each instance and algorithm, and then reported the average values.

\subsection{Summary of the results} \label{sec:computational_experiments}

We calculated the four metrics in each algorithm (\emph{ceyhan}, \emph{holzmann}, {\AMOCO}),  for a fixed time in the set $\{10,60,300,900\}$, measured in seconds. 
We have chosen this set of cut-times to know the performance of each algorithm at the very beginning of the run (just 10 s), and also after a longer (but still short) time (900 s), plus some other intermediate times. 

Tables~\ref{tab1:frecuencies} and \ref{tab1:frecuencies2} present the results for each algorithm and metric. There are two columns per combination. The first one contains the number of instances in which the algorithm has the best average value (sometimes shared with another algorithm). The second column contains the number of instances in which the algorithm is the only one obtaining the best average value. The results are displayed for the three classes of the benchmark (\emph{AP}, \emph{ILP} and \emph{KP}) at four cut-points each, and the total sum is shown in the four last rows of the table.

\begin{table*}[h] 
\scriptsize
\centering
\setlength{\tabcolsep}{3pt}
\begin{tabular}{clrrcrrcrrcrrcrrcrr}
& &  \multicolumn{8}{c}{ONVGR} & & \multicolumn{8}{c}{HV} \\
\cmidrule{3-10}
\cmidrule{12-19}
& &  \multicolumn{2}{c}{\emph{ceyhan}} & & \multicolumn{2}{c}{\emph{holzmann}} & &  \multicolumn{2}{c}{\AMOCO} & &
  \multicolumn{2}{c}{\emph{ceyhan}} & & \multicolumn{2}{c}{\emph{holzmann}} & & \multicolumn{2}{c}{\AMOCO}  \tabularnewline
\cmidrule{3-4}
\cmidrule{6-7}
\cmidrule{9-10}
\cmidrule{12-13}
\cmidrule{15-16}
\cmidrule{18-19}
& & 
 \multicolumn{1}{c}{\emph{best}} & \multicolumn{1}{c}{\emph{excl}} & &
 \multicolumn{1}{c}{\emph{best}} & \multicolumn{1}{c}{\emph{excl}} & &
 \multicolumn{1}{c}{\emph{best}} & \multicolumn{1}{c}{\emph{excl}} & &  
 \multicolumn{1}{c}{\emph{best}} & \multicolumn{1}{c}{\emph{excl}} & &
 \multicolumn{1}{c}{\emph{best}} & \multicolumn{1}{c}{\emph{excl}} & &
 \multicolumn{1}{c}{\emph{best}} & \multicolumn{1}{c}{\emph{excl}}  \tabularnewline
  
\toprule
\multirow{4}{*} {\emph{AP}}  
& $t$=\phantom{0}10 s  & 11 & 0 & & 34 & 21 & & 79 & 66 & & 11 & 0 & & 45 & 32 & & 68  & 55 \\
& $t$=\phantom{0}60 s  & 20 & 0 & & 31 & 8  & & 92 & 69 & & 20 & 0 & & 32 & 9  & & 91  & 68 \\
& $t$=300 s            & 29 & 0 & & 66 & 33 & & 67 & 34 & & 29 & 0 & & 33 & 0  & & 100 & 67 \\
& $t$=900 s            & 33 & 0 & & 52 & 7  & & 93 & 48 & & 33 & 0 & & 45 & 0  & & 100 & 55 \tabularnewline
\midrule
\multirow{4}{*} {\emph{ILP}}  
& $t$=\phantom{0}10 s  & 44 & 8 & & 116 & 55 & & 157 & 96 & & 43 & 7 & & 119 & 58 & & 155 & 94 \\
& $t$=\phantom{0}60 s  & 51 & 3 & & 134 & 48 & & 169 & 83 & & 50 & 2 & & 120 & 33 & & 185 & 98 \\
& $t$=300 s            & 61 & 1 & & 146 & 35 & & 184 & 73 & & 60 & 0 & & 129 & 18 & & 202 & 91 \\
& $t$=900 s            & 76 & 0 & & 158 & 19 & & 201 & 62 & & 76 & 0 & & 146 & 7  & & 213 & 74 \tabularnewline
\midrule
\multirow{4}{*} {\emph{KP}}  
& $t$=\phantom{0}10 s  & 41 & 0 & & 79  & 22 & & 138 & 81 & & 41 & 0 & & 95 & 38 & & 122 & 65 \\
& $t$=\phantom{0}60 s  & 56 & 0 & & 97  & 4  & & 156 & 63 & & 56 & 0 & & 93 & 1  & & 159 & 67 \\
& $t$=300 s            & 71 & 0 & & 128 & 0  & & 160 & 32 & & 71 & 0 & & 127 & 0 & & 159 & 33 \\
& $t$=900 s            & 80 & 0 & & 149 & 0  & & 160 & 11 & & 80 & 0 & & 148 & 3 & & 156 & 12 \tabularnewline
\midrule
\midrule
\multirow{4}{*} {TOTAL}  
& $t$=\phantom{0}10 s  & 96  & 8 & & 229 & 98 & & 374 & 243 & & 95  & 7 & & 259 & 128 & & 345 & 214 \\
& $t$=\phantom{0}60 s  & 127 & 3 & & 262 & 60 & & 417 & 215 & & 126 & 2 & & 245 & 43  & & 435 & 233 \\
& $t$=300 s            & 161 & 1 & & 340 & 68 & & 411 & 139 & & 160 & 0 & & 289 & 18  & & 461 & 191 \\
& $t$=900 s            & 189 & 0 & & 359 & 26 & & 454 & 121 & & 189 & 0 & & 339 & 10  & & 469 & 141 \tabularnewline
\bottomrule
\end{tabular}

\caption{Number of times that each algorithm obtains the best average value without exclusivity (\emph{best}) and with exclusivity (\emph{excl}), for ONVGR and HV, for different execution times, and for each class in the benchmark. Total sum is displayed in the last four rows.}
\label{tab1:frecuencies}
\end{table*}

\begin{table*}[h] 
\scriptsize
\centering
\setlength{\tabcolsep}{3pt}
\begin{tabular}{clrrcrrcrrcrrcrrcrr}
& &  \multicolumn{8}{c}{$\Delta^{*}$} & & \multicolumn{8}{c}{$\varepsilon_{+}$} \\
\cmidrule{3-10}
\cmidrule{12-19}
& &  \multicolumn{2}{c}{\emph{ceyhan}} & & \multicolumn{2}{c}{\emph{holzmann}} & &  \multicolumn{2}{c}{\AMOCO} & &
  \multicolumn{2}{c}{\emph{ceyhan}} & & \multicolumn{2}{c}{\emph{holzmann}} & & \multicolumn{2}{c}{\AMOCO}  \tabularnewline
\cmidrule{3-4}
\cmidrule{6-7}
\cmidrule{9-10}
\cmidrule{12-13}
\cmidrule{15-16}
\cmidrule{18-19}
& & 
 \multicolumn{1}{c}{\emph{best}} & \multicolumn{1}{c}{\emph{excl}} & &
 \multicolumn{1}{c}{\emph{best}} & \multicolumn{1}{c}{\emph{excl}} & &
 \multicolumn{1}{c}{\emph{best}} & \multicolumn{1}{c}{\emph{excl}} & &  
 \multicolumn{1}{c}{\emph{best}} & \multicolumn{1}{c}{\emph{excl}} & &
 \multicolumn{1}{c}{\emph{best}} & \multicolumn{1}{c}{\emph{excl}} & &
 \multicolumn{1}{c}{\emph{best}} & \multicolumn{1}{c}{\emph{excl}}  \tabularnewline

\toprule
\multirow{4}{*} {\emph{AP}}  
& $t$=\phantom{0}10 s  & 75 & 64 & & 29 & 17 & & 19 & 7  & & 11 & 0 & & 27 & 14 & & 86  & 73 \\
& $t$=\phantom{0}60 s  & 59 & 39 & & 40 & 20 & & 41 & 21 & & 20 & 0 & & 25 & 2  & & 98  & 75 \\
& $t$=300 s            & 44 & 15 & & 43 & 13 & & 72 & 42 & & 29 & 0 & & 34 & 1  & & 99  & 66 \\
& $t$=900 s            & 54 & 21 & & 36 & 3  & & 76 & 43 & & 33 & 0 & & 45 & 0  & & 100 & 55   \tabularnewline
\midrule
\multirow{4}{*} {\emph{ILP}}  
& $t$=\phantom{0}10 s  & 143 & 107 & & 75  & 27 & & 86  & 38 & & 44 & 8 & & 133 & 68 & & 144 & 79 \\ 
& $t$=\phantom{0}60 s  & 163 & 115 & & 81  & 21 & & 84  & 24 & & 51 & 3 & & 137 & 44 & & 173 & 80 \\
& $t$=300 s            & 168 & 108 & & 87  & 11 & & 101 & 25 & & 61 & 1 & & 144 & 32 & & 187 & 75 \\
& $t$=900 s            & 179 & 103 & & 100 & 6  & & 111 & 17 & & 76 & 0 & & 160 & 21 & & 199 & 60   \tabularnewline
\midrule
\multirow{4}{*} {\emph{KP}}  
& $t$=\phantom{0}10 s  & 121 & 80 & & 63 & 14 & & 66 & 17 & & 42 & 1 & & 94  & 37 & & 122 & 65 \\
& $t$=\phantom{0}60 s  & 123 & 67 & & 70 & 3  & & 90 & 23 & & 56 & 0 & & 97  & 5  & & 155 & 63 \\ 
& $t$=300 s            & 139 & 69 & & 86 & 2  & & 89 & 5  & & 71 & 0 & & 127 & 0  & & 159 & 33 \\ 
& $t$=900 s            & 145 & 66 & & 93 & 2  & & 92 & 1  & & 80 & 0 & & 148 & 3  & & 156 & 12 \tabularnewline
\midrule
\midrule
\multirow{4}{*} {TOTAL}  
& $t$=\phantom{0}10 s  & 339 & 251 & & 167 & 58 & & 171 & 62 & & 97  & 9 & & 254 & 119 & & 352 & 217 \\
& $t$=\phantom{0}60 s  & 345 & 221 & & 191 & 44 & & 215 & 68 & & 127 & 3 & & 259 & 51  & & 426 & 218 \\
& $t$=300 s            & 351 & 192 & & 216 & 26 & & 262 & 72 & & 161 & 1 & & 305 & 33  & & 445 & 174 \\
& $t$=900 s            & 378 & 190 & & 229 & 11 & & 279 & 61 & & 189 & 0 & & 353 & 24  & & 455 & 127  \tabularnewline
\bottomrule
\end{tabular}

\caption{Number of times that each algorithm obtains the best average value without exclusivity (\emph{best}) and with exclusivity (\emph{excl}), for $\Delta^{*}$ and $\varepsilon_{+}$, for different execution times, and for each class in the benchmark. Total sum is displayed in the last four rows.}

\label{tab1:frecuencies2}
\end{table*}

Analyzing the ONVGR metric, we can see that \emph{holzmann} and {\AMOCO} provide the higher number of solutions. In fact, {\AMOCO} is better in all the classes in the benchmark. If we observe the row with the total sum of the three classes, the new proposed algorithm is clearly the best.
Sometimes the number of solutions is not the most important issue. In fact, the hypervolume is considered one of the best metrics to measure the spread over the objective space. Looking at the results for HV, we see that in all the classes the total hypervolume reached by {\AMOCO} is also maximum. It is the clear winner, with a large difference with respect to the second-best algorithm, \emph{holzmann}. Note that \emph{ceyhan} has the best hypervolume in a few cases. For instance, in the \emph{ILP} class, at 10 seconds, \emph{ceyhan} has the best value 43 times, but only in 7 cases it is the unique winner. In the other 36 instances, its HV is similar to that of other algorithms. The great differences in HV may be explained because, overall, \emph{ceyhan} finds much fewer solutions than the others and, consequently, a worse total hypervolume. \emph{Ceyhan} is good at finding a few solutions quickly, with a very good general spread, as we can see in the results for $\Delta^{*}$ in Table~\ref{tab1:frecuencies2}, where it is the clear winner. For the last metric, $\varepsilon_{+}$, Table~\ref{tab1:frecuencies2} shows again that {\AMOCO} has the best results in all cases.

As a summary of the results shown in Tables~\ref{tab1:frecuencies} and \ref{tab1:frecuencies2}, we see that \emph{holzmann} and {\AMOCO} provide better performance than \emph{ceyhan}. This means that the structure of the algorithm has an influence on the final results. 
While \emph{ceyhan} is designed to find a few well-spread solutions quickly, the dispersion of the solutions in the objective space is not so good when the search progresses.  \emph{Ceyhan} makes more than one call to the ILP solver at each iteration because it uses a subset of vectors which depends on the number of solutions found so far. We noticed in the experiments that these multiple calls to the solver per iteration seem to degrade performance as the search progresses.
The other two methods (\emph{holzmann} and {\AMOCO}), based on the framework described in Algorithm~\ref{alg:general_MOCO}, seem to be more efficient for longer run execution times. There is an exception in the results for $\Delta^{*}$, where \emph{ceyhan} is the clear winner. 
Comparing \emph{holzmann} and {\AMOCO} methods, the latter obtains better results. Although these two algorithms are based on the same framework and use the same mathematical program, {\AMOCO} changes the way in which the next boxes to explore are selected: by alternating directions, using \emph{p-partition}, and using a different priority value assigned to each box. The results demonstrate that the three new contributions described in this paper have a positive effect on the anytime performance.

\subsection{Detailed analysis} \label{sec:performance_analysis}
This subsection is divided into three third-level sections. 
In the first third-level section, we rank the algorithms according to their performance on the instances. Clarifying graphs are shown at ten regular cut-points from~90 s to 900~s. In the second third-level section, we show the evolution of the metrics over time for three selected instances. In this way, we observe the anytime performance of the algorithms for each metric. The last third-level section provides the results of non-parametric statistical tests to check if the differences are statistically significant or not. In the supplementary material attached to this paper, we group the instances into categories, and show tables with the results for the four metrics at four time points.

\subsubsection{Ranking the algorithms in each class of instances}

To better measure the performance of the algorithms, we computed a rank for each of them in each instance.
We did this for ONVGR, HVR, $\Delta^{*}$, and $\varepsilon_{+}$, at different stopping times from~90 s to 900~s, with a step of 90~s. For example, fixing a time limit of 360~s, we compared the average ONVGR value of the 30 runs of the algorithms in the first instance and assigned a rank from~1 to~3, being~1 the algorithm with the best value for the metric. We did the same in the rest of the cases. In the case of ties, we proceeded by taking the average rankings for the ties (e.g., 1.5 for a tie between the first two algorithms).

\begin{figure} [ht]
\centering
\begin{minipage} {0.6\linewidth}
\begin{center}
\includegraphics[scale=0.22]{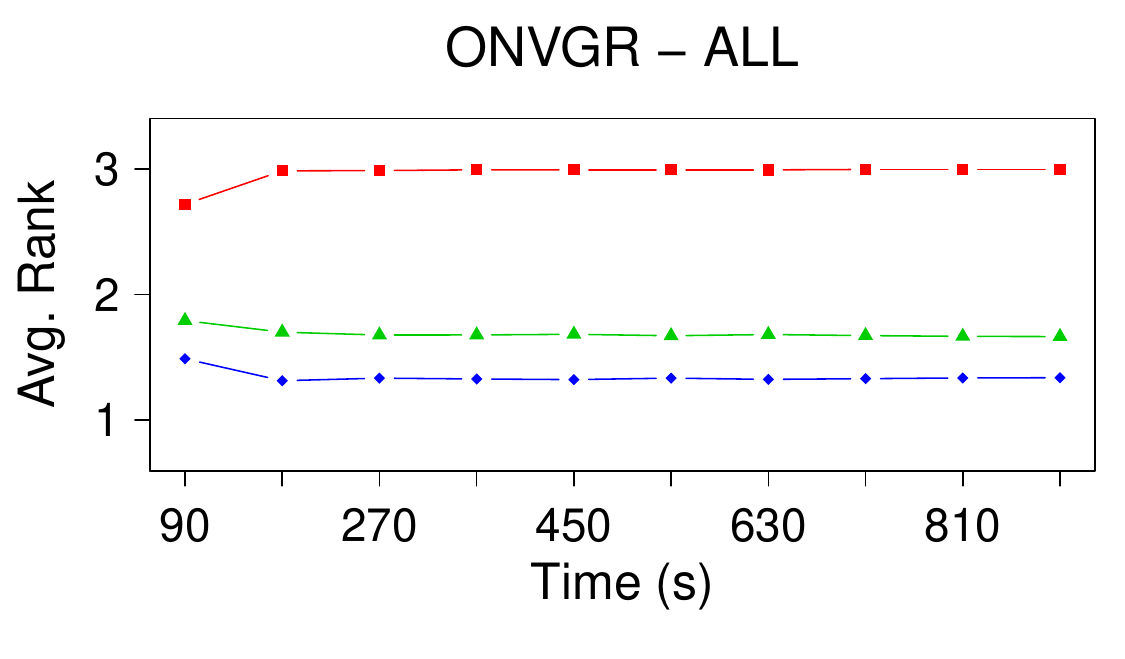}
\end{center}
\end{minipage}
\begin{minipage} {0.2\linewidth}
\includegraphics[scale=0.2]{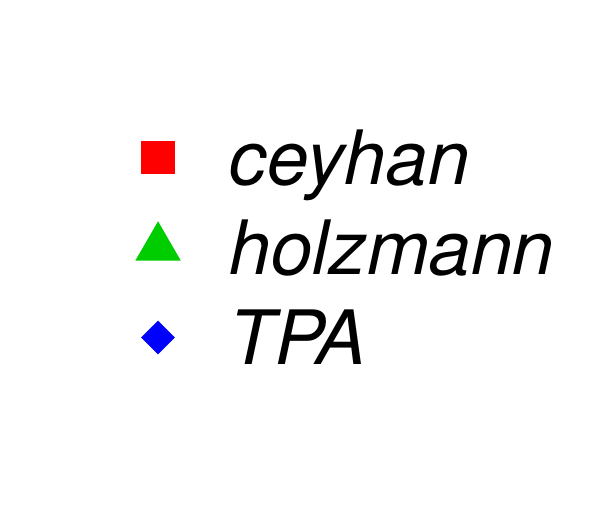}
\end{minipage}
\begin{minipage} {0.325\linewidth}
\includegraphics[scale=0.215]{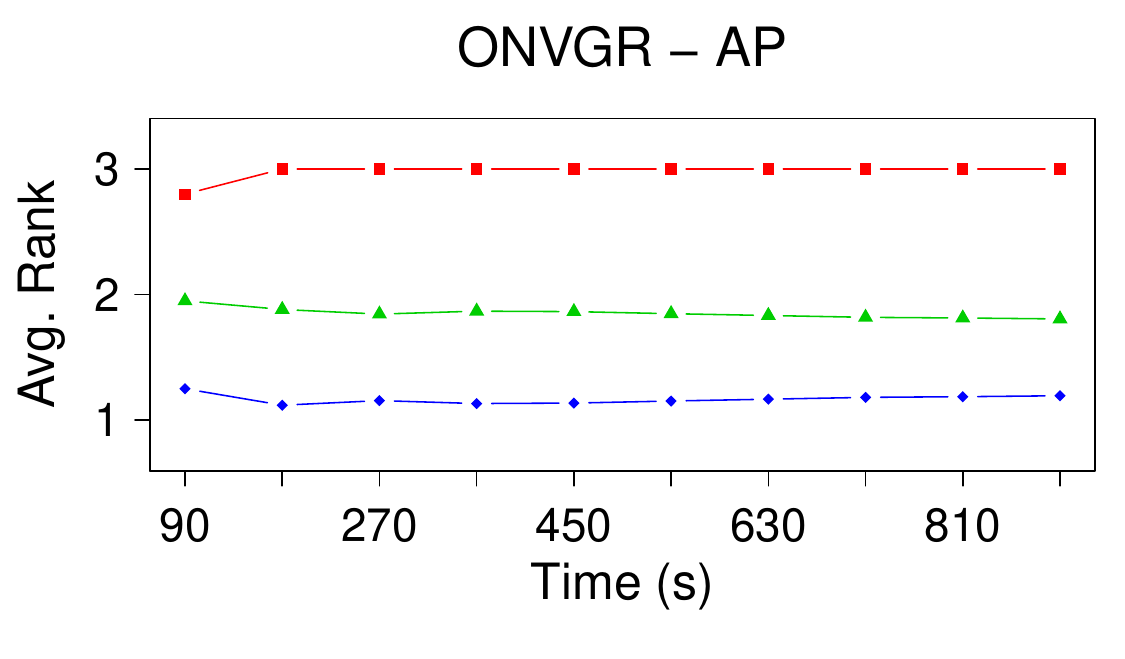}
\end{minipage}
\begin{minipage} {0.325\linewidth}
\includegraphics[scale=0.215]{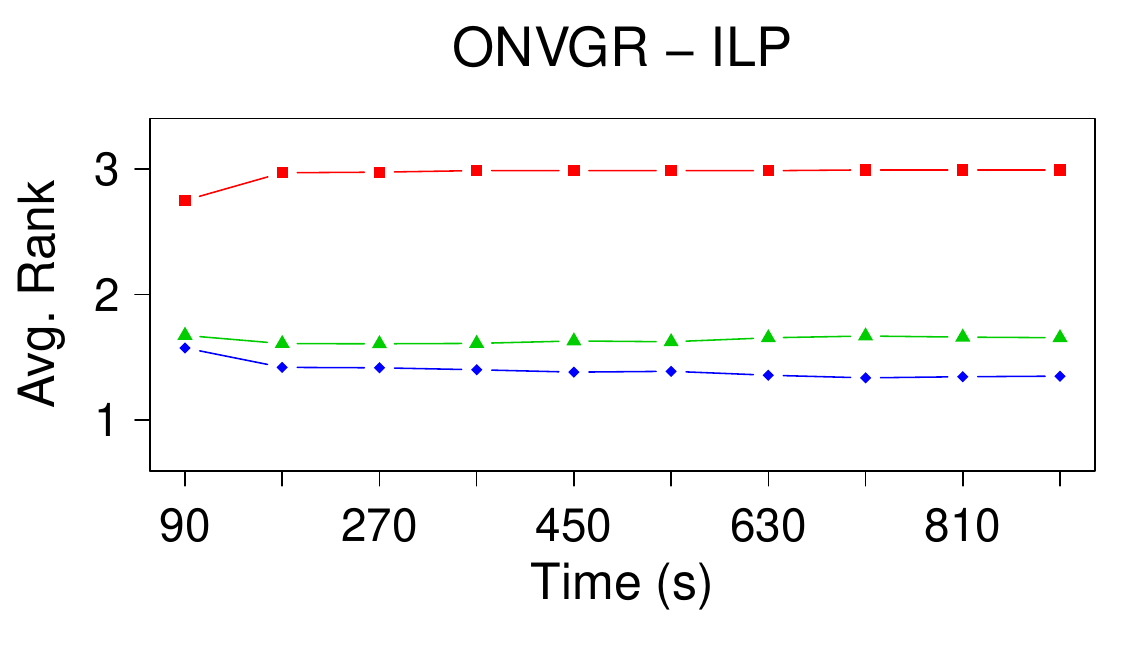}
\end{minipage}
\begin{minipage} {0.325\linewidth}
\includegraphics[scale=0.215]{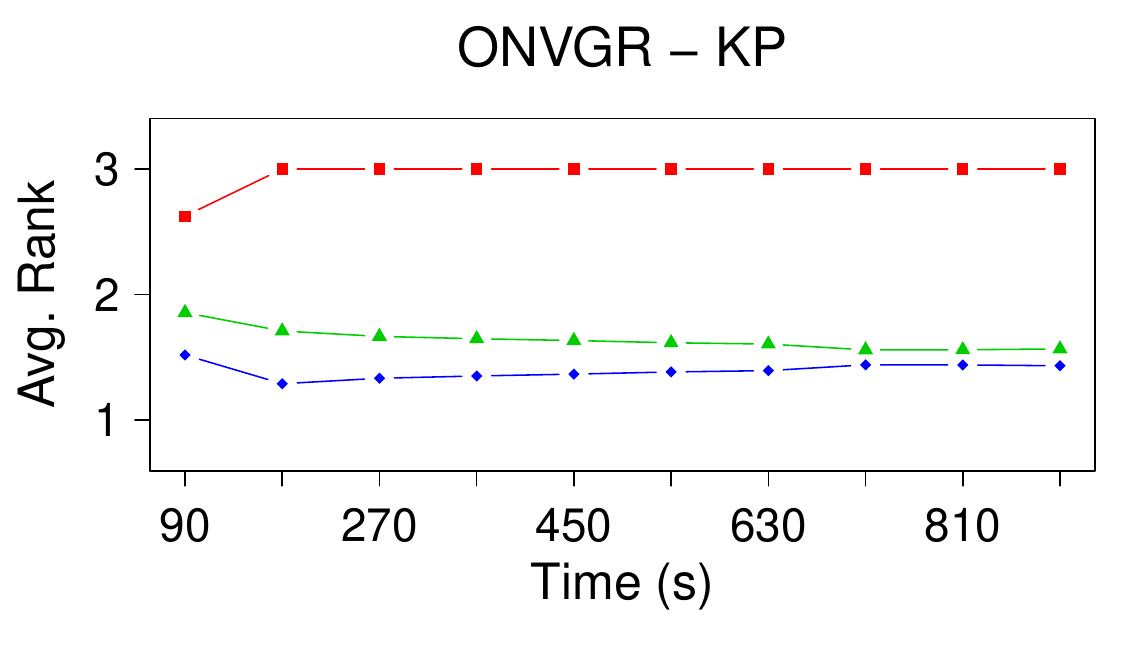}
\end{minipage}

\begin{minipage} {0.6\linewidth}
\begin{center}
\includegraphics[scale=0.22]{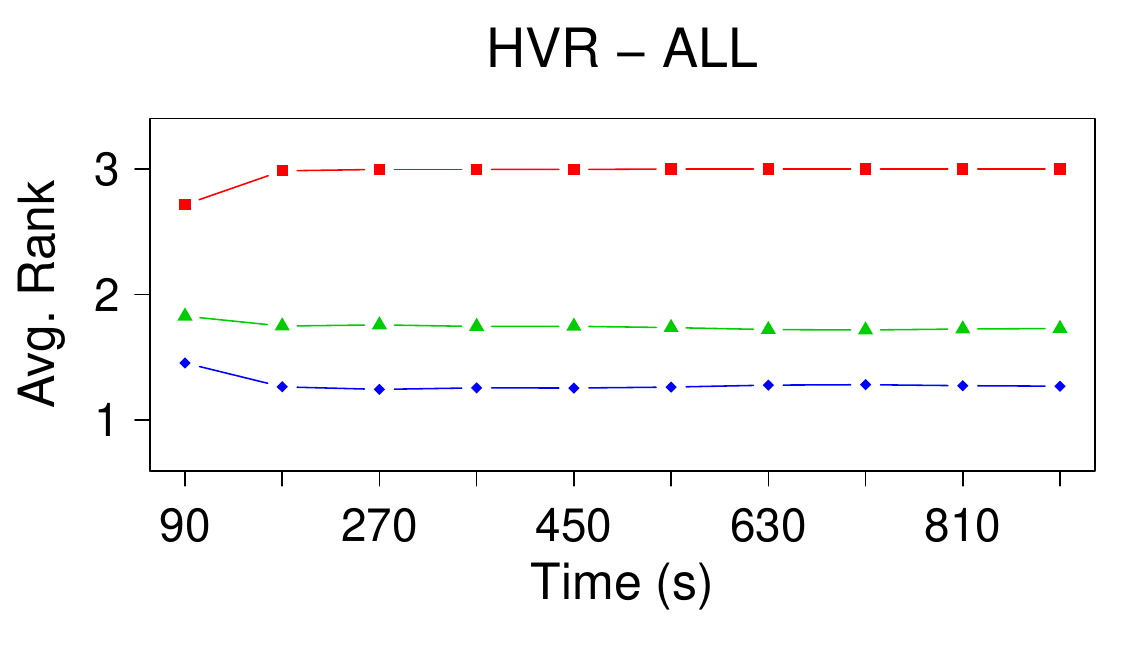}
\end{center}
\end{minipage}
\begin{minipage} {0.2\linewidth}
\includegraphics[scale=0.2]{images/vertical_legend.pdf}
\end{minipage}
\begin{minipage} {0.325\linewidth}
\includegraphics[scale=0.215]{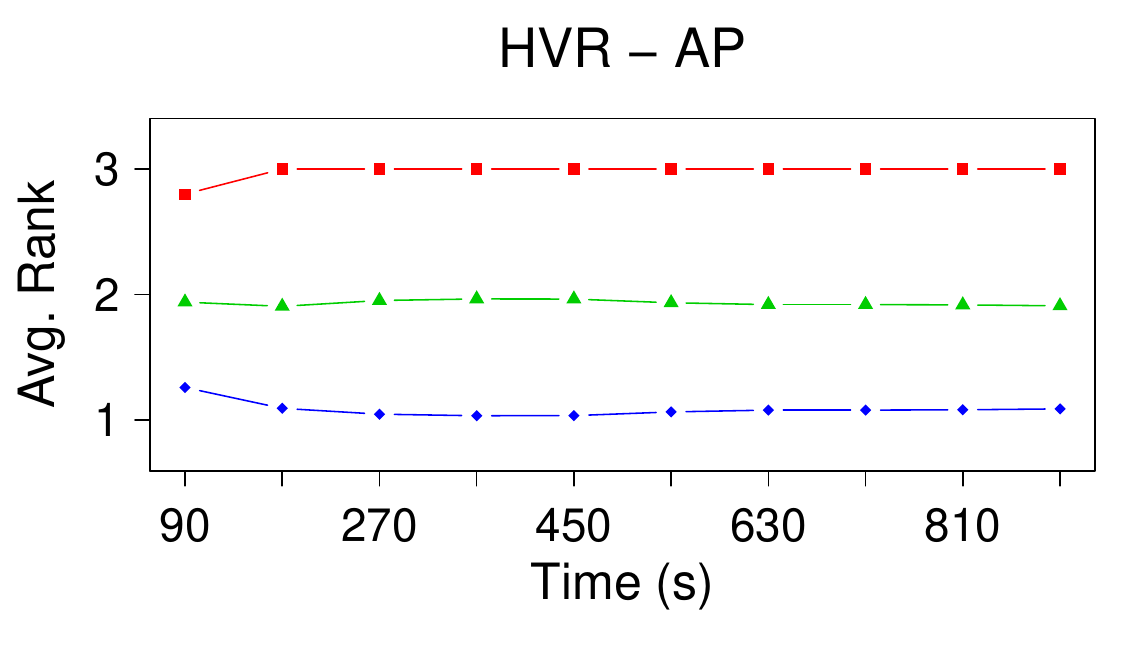}
\end{minipage}
\begin{minipage} {0.325\linewidth}
\includegraphics[scale=0.215]{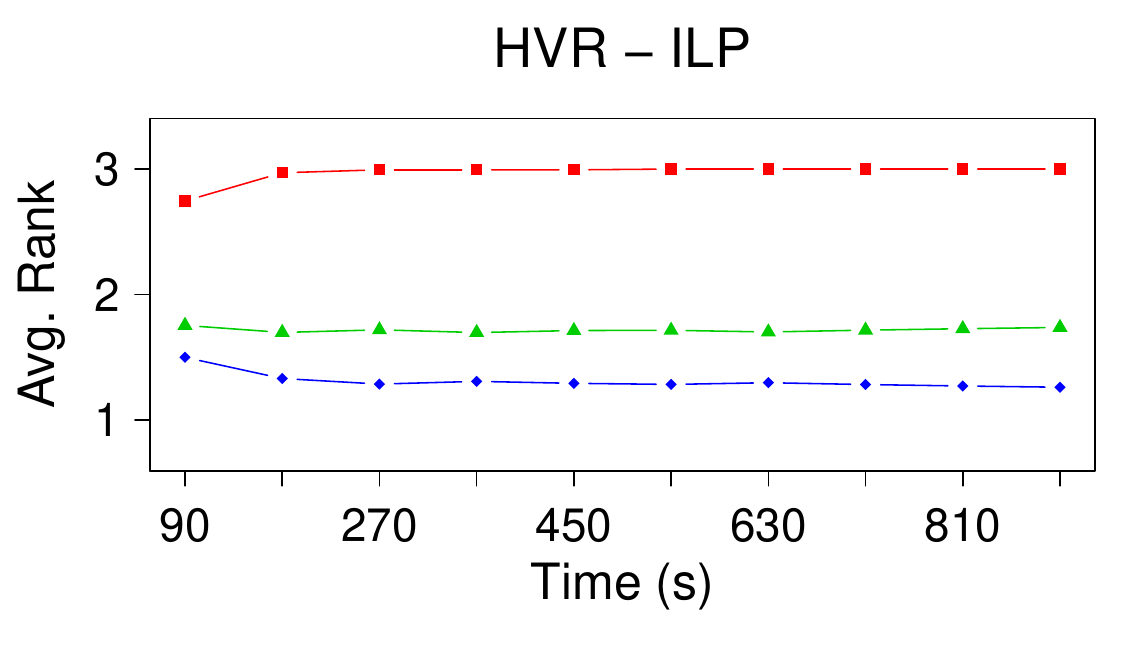}
\end{minipage}
\begin{minipage} {0.325\linewidth}
\includegraphics[scale=0.215]{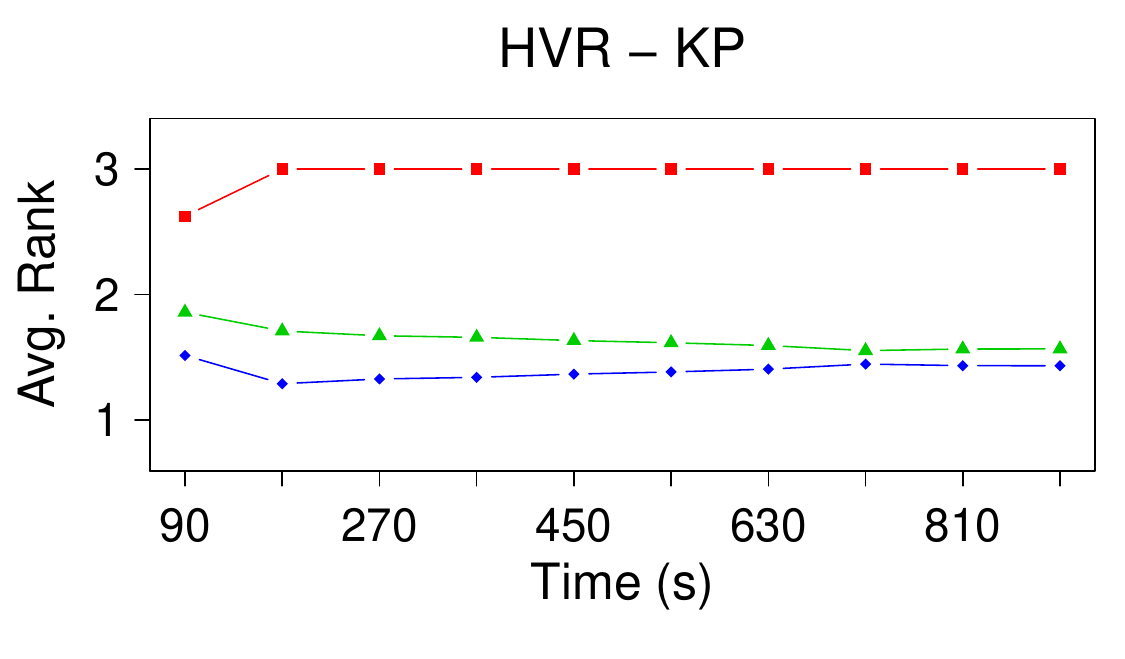}
\end{minipage}

\caption{Average rank values for the algorithms at 10 cut-points, using the metrics ONVGR and HVR. The lower the rank, the better the algorithm.} \label{fig:rank-1}
\end{figure}

\begin{figure} [ht]
\centering
\begin{minipage} {0.6\linewidth}
\begin{center}
\includegraphics[scale=0.22]{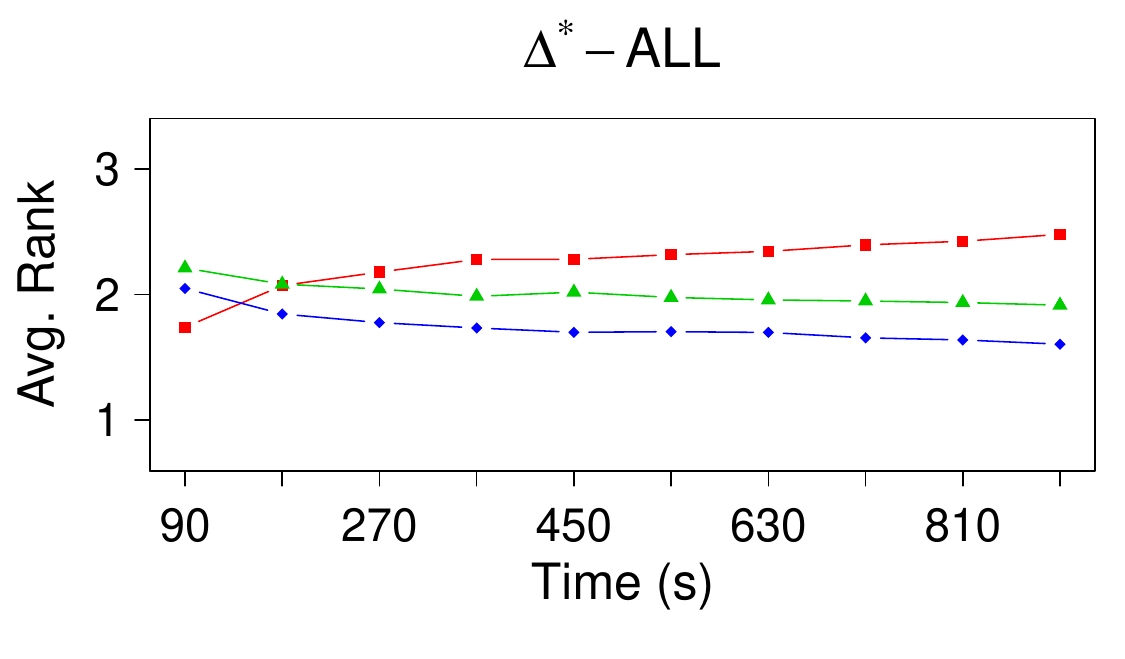}
\end{center}
\end{minipage}
\begin{minipage} {0.2\linewidth}
\includegraphics[scale=0.2]{images/vertical_legend.pdf}
\end{minipage}
\begin{minipage} {0.325\linewidth}
\includegraphics[scale=0.215]{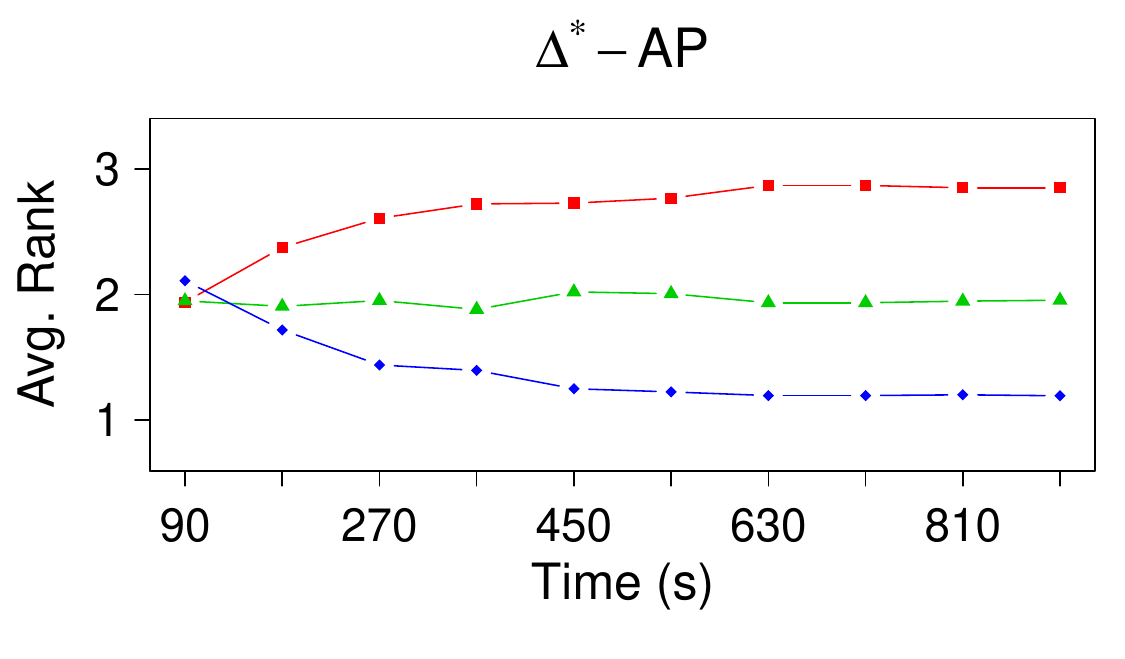}
\end{minipage}
\begin{minipage} {0.325\linewidth}
\includegraphics[scale=0.215]{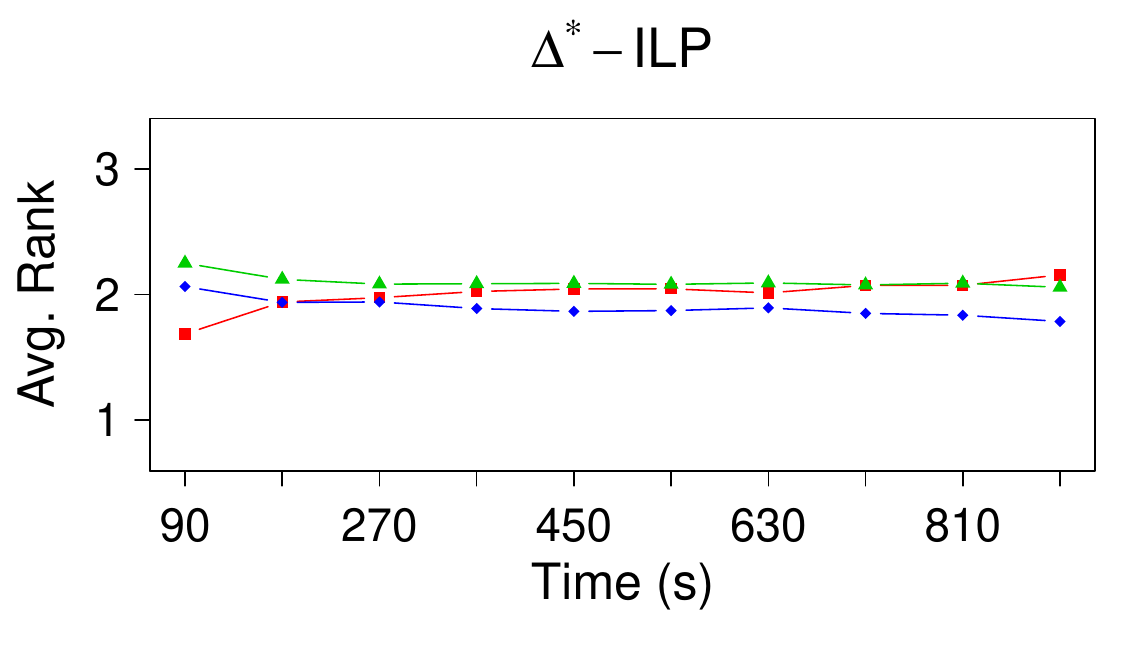}
\end{minipage}
\begin{minipage} {0.325\linewidth}
\includegraphics[scale=0.215]{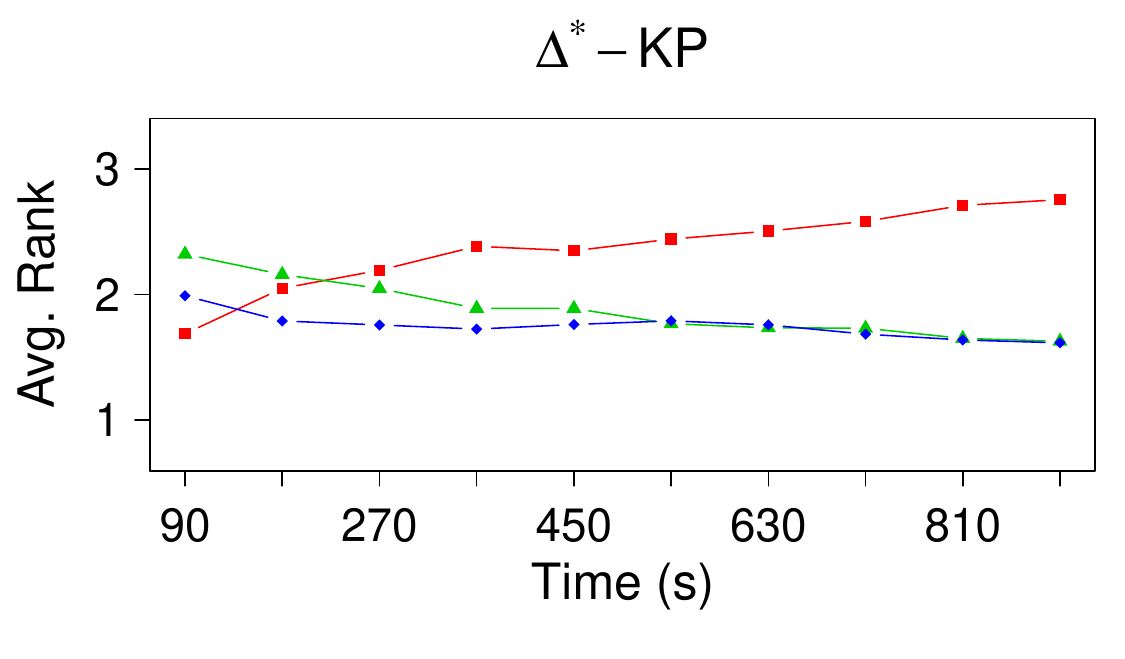}
\end{minipage}

\begin{minipage} {0.6\linewidth}
\begin{center}
\includegraphics[scale=0.22]{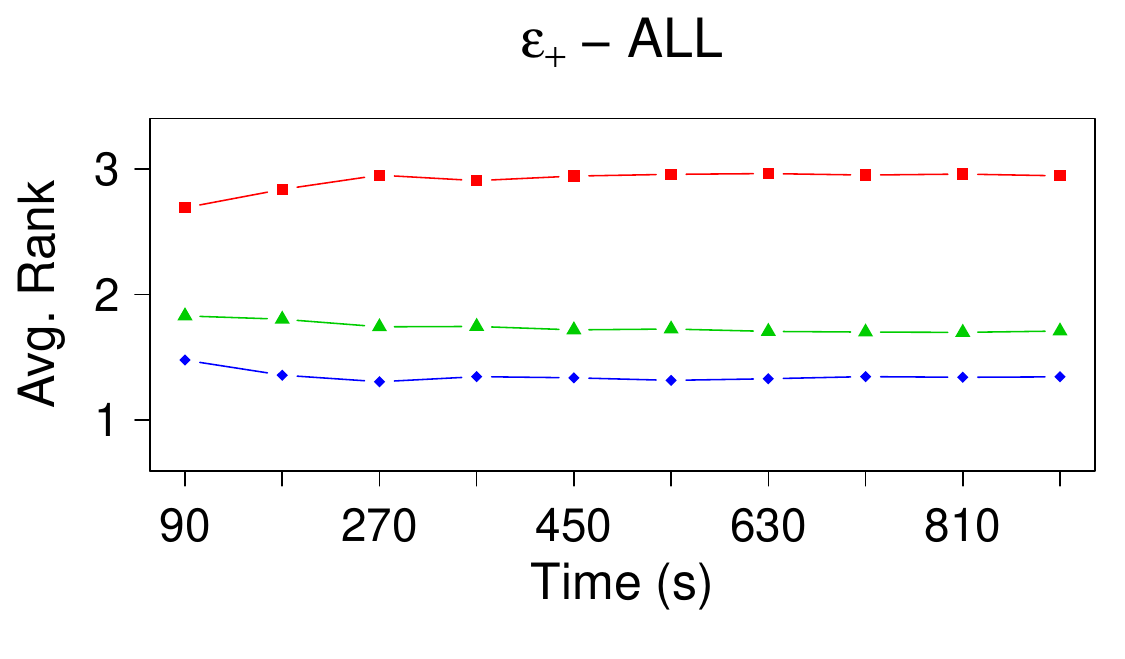}
\end{center}
\end{minipage}
\begin{minipage} {0.2\linewidth}
\includegraphics[scale=0.2]{images/vertical_legend.pdf}
\end{minipage}
\begin{minipage} {0.325\linewidth}
\includegraphics[scale=0.215]{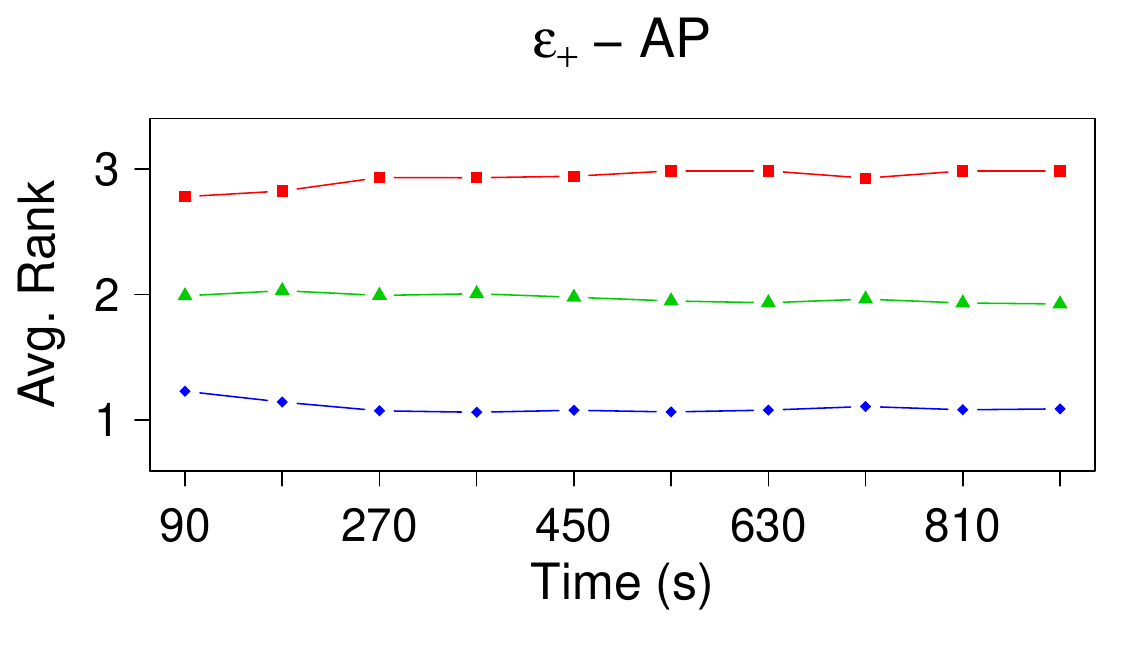}
\end{minipage}
\begin{minipage} {0.325\linewidth}
\includegraphics[scale=0.215]{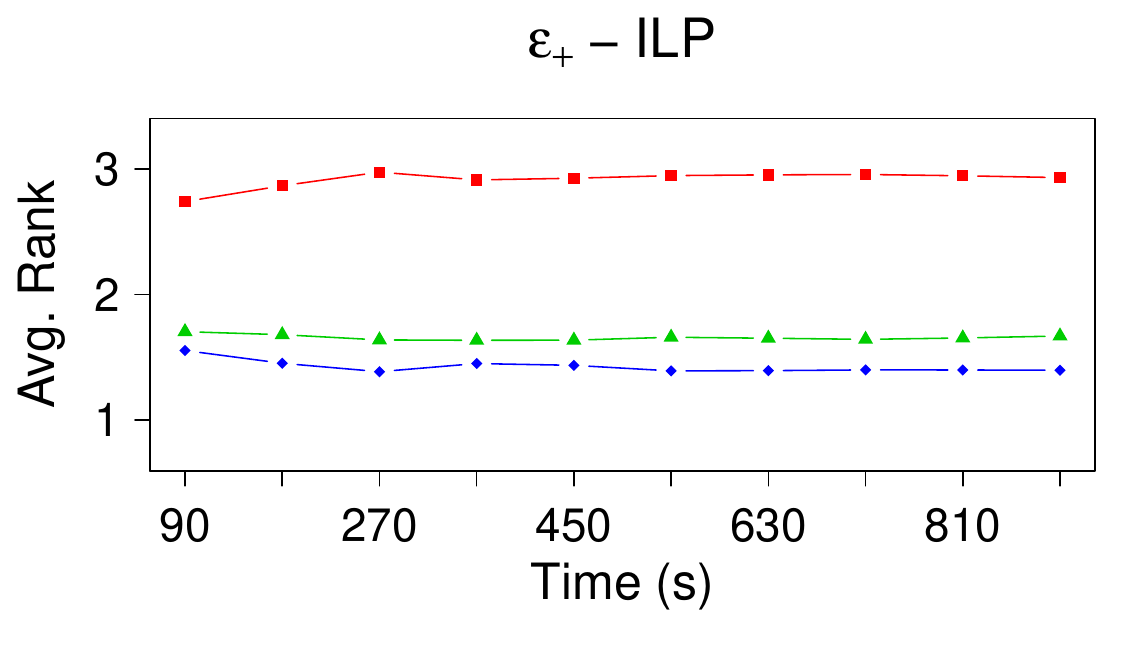}
\end{minipage}
\begin{minipage} {0.325\linewidth}
\includegraphics[scale=0.215]{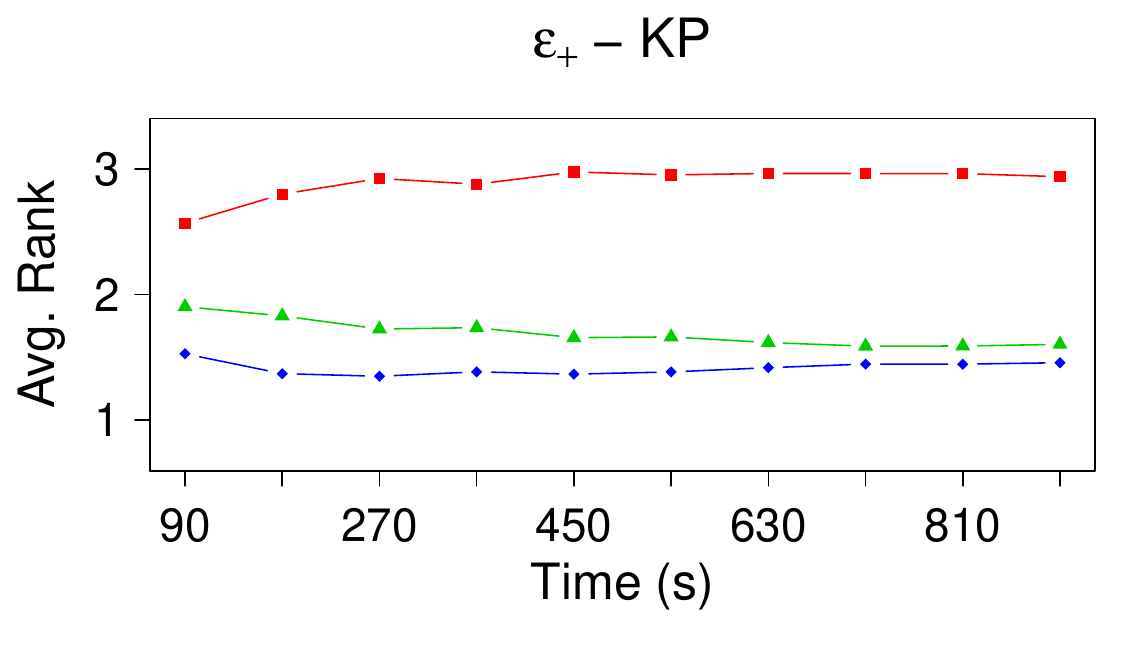}
\end{minipage}

\caption{Average rank values for the algorithms at 10 cut-points, using the metrics $\Delta^{*}$ and $\varepsilon_{+}$. The lower the rank, the better the algorithm.} \label{fig:rank-2}
\end{figure}

There are some time points in which an algorithm may have finished its execution and others not. In those cases, a rank value of 1 is assigned to the algorithm that finished and the following rankings are distributed among the rest. When all the algorithms have finished at a given time in all the runs, the instance is discarded. 
The results of Figures~\ref{fig:rank-1} and~\ref{fig:rank-2} indicate that the best average rank values correspond to {\AMOCO} in most of the cases with important differences compared to the other metrics in some cases.

\subsubsection{Search progress}
In this section we want to show the progress of the search on some selected instances, to further illustrate how the algorithms evolve and behave at any time. For each class of the benchmark, the instance which requires the longest time to be solved is selected. 

\begin{figure} [t]
\begin{center}
\begin{minipage} {0.326\linewidth}
\includegraphics[scale=0.135]{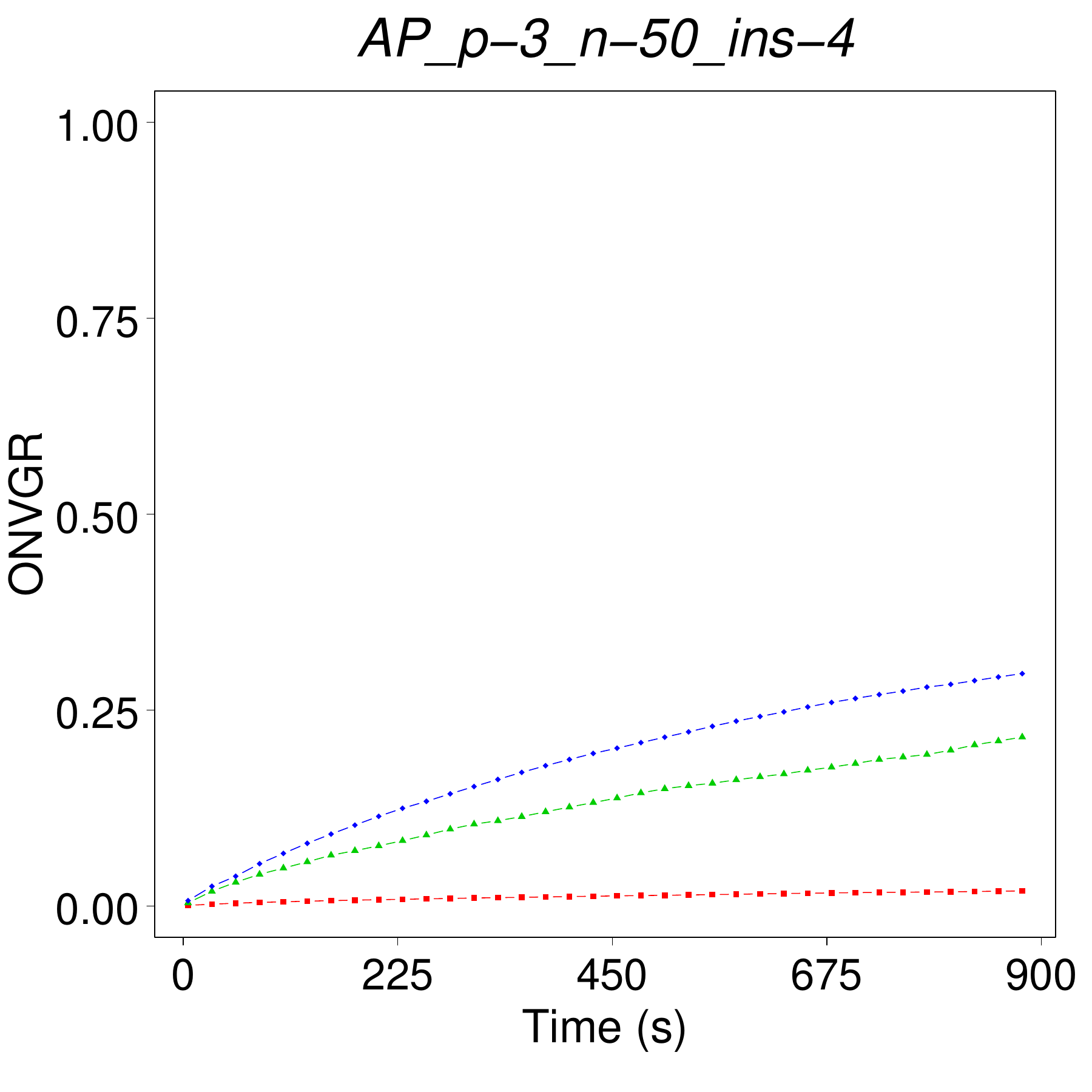}
\end{minipage}
\begin{minipage} {0.326\linewidth}
\includegraphics[scale=0.135]{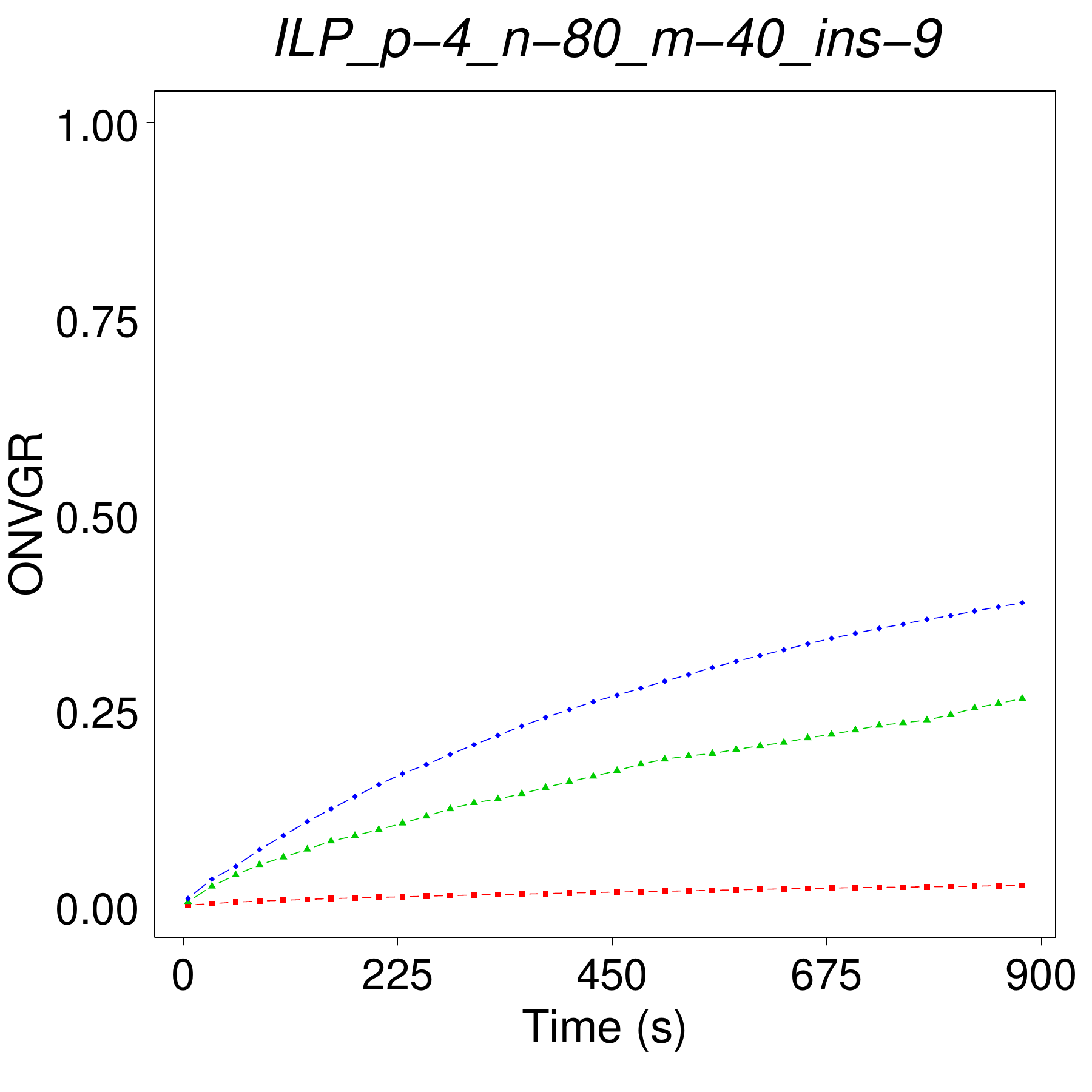} 
\end{minipage} 
\begin{minipage} {0.326\textwidth}
\includegraphics[scale=0.135]{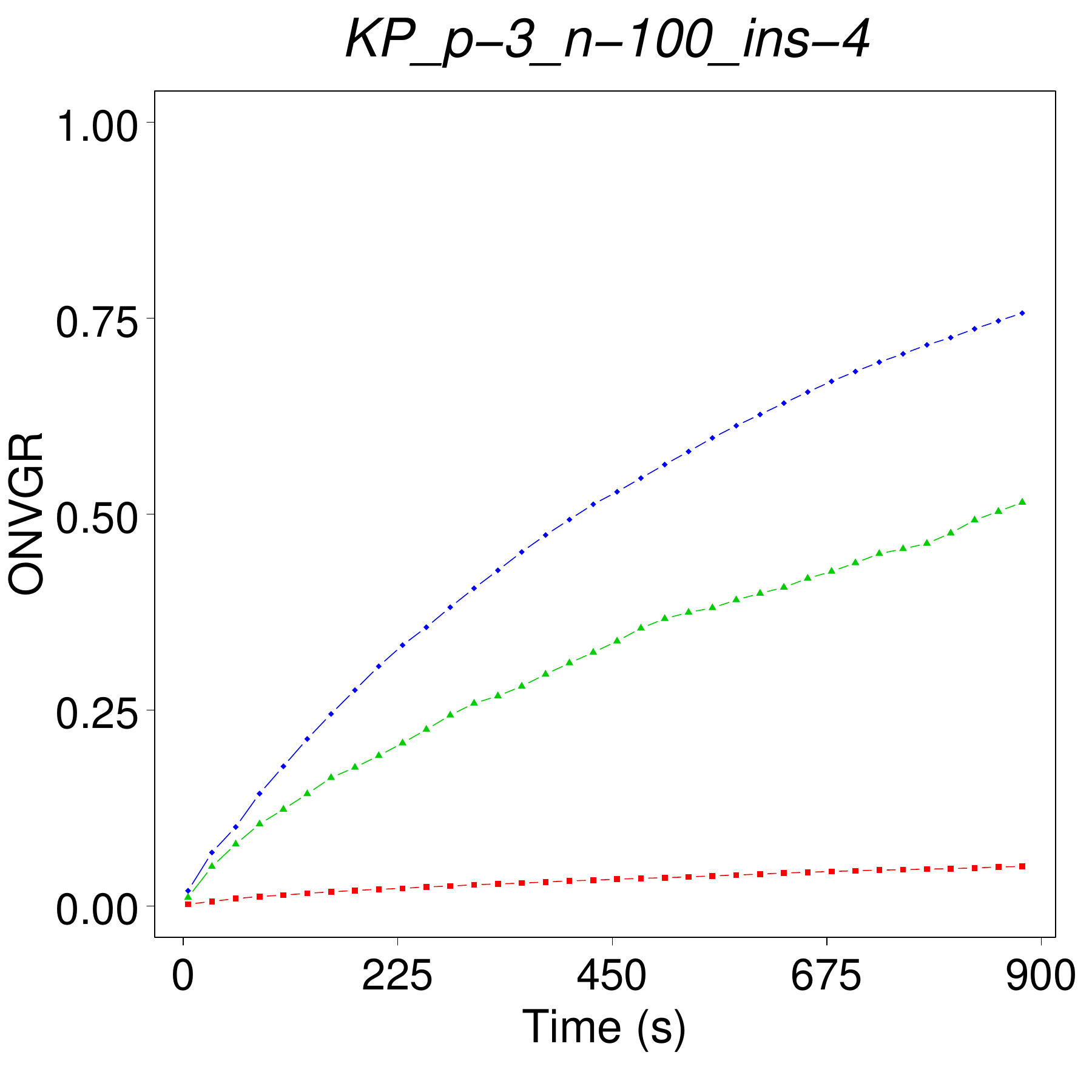} 
\end{minipage}

\begin{minipage} {0.326\linewidth}
\includegraphics[scale=0.135]{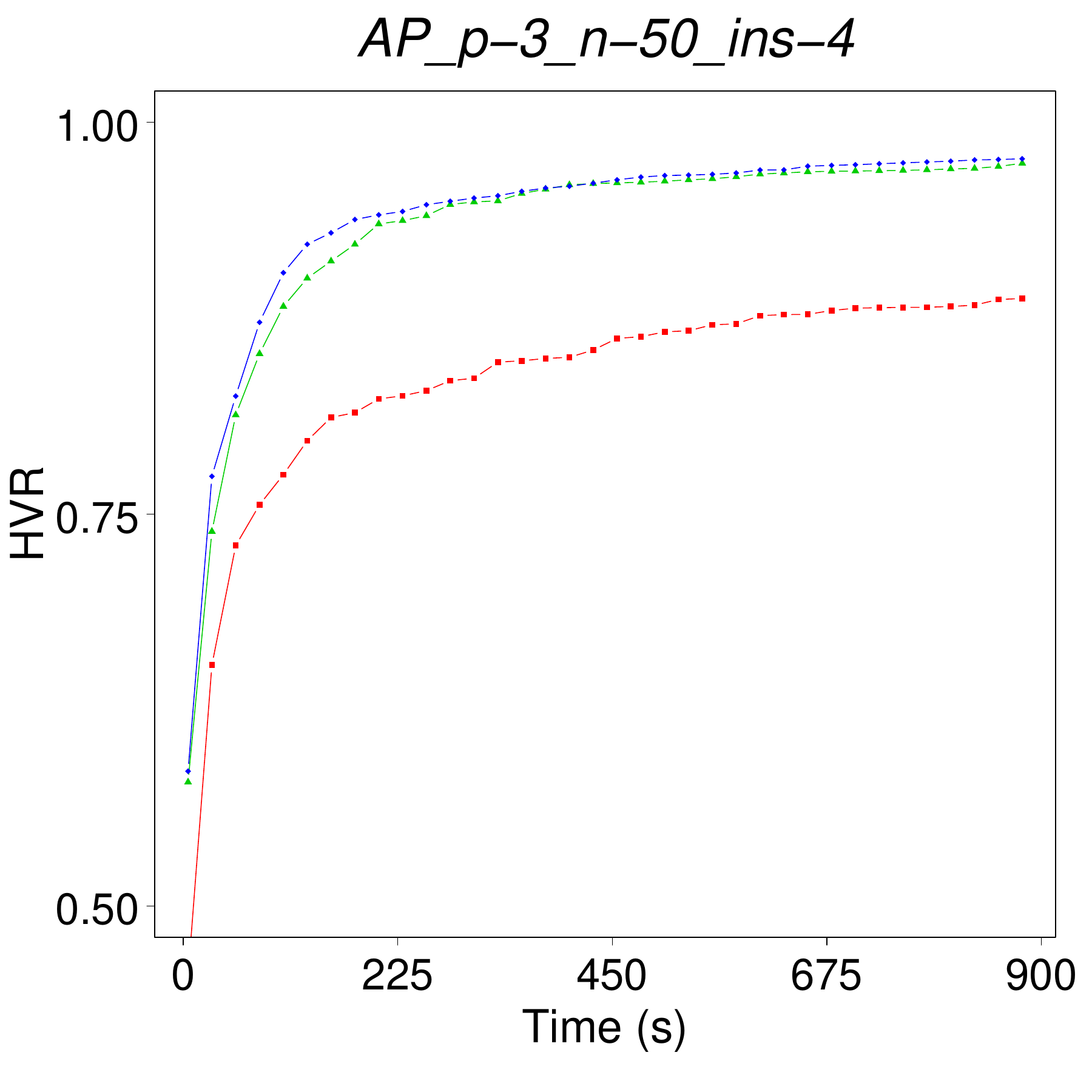}
\end{minipage}
\begin{minipage} {0.326\linewidth}
\includegraphics[scale=0.135]{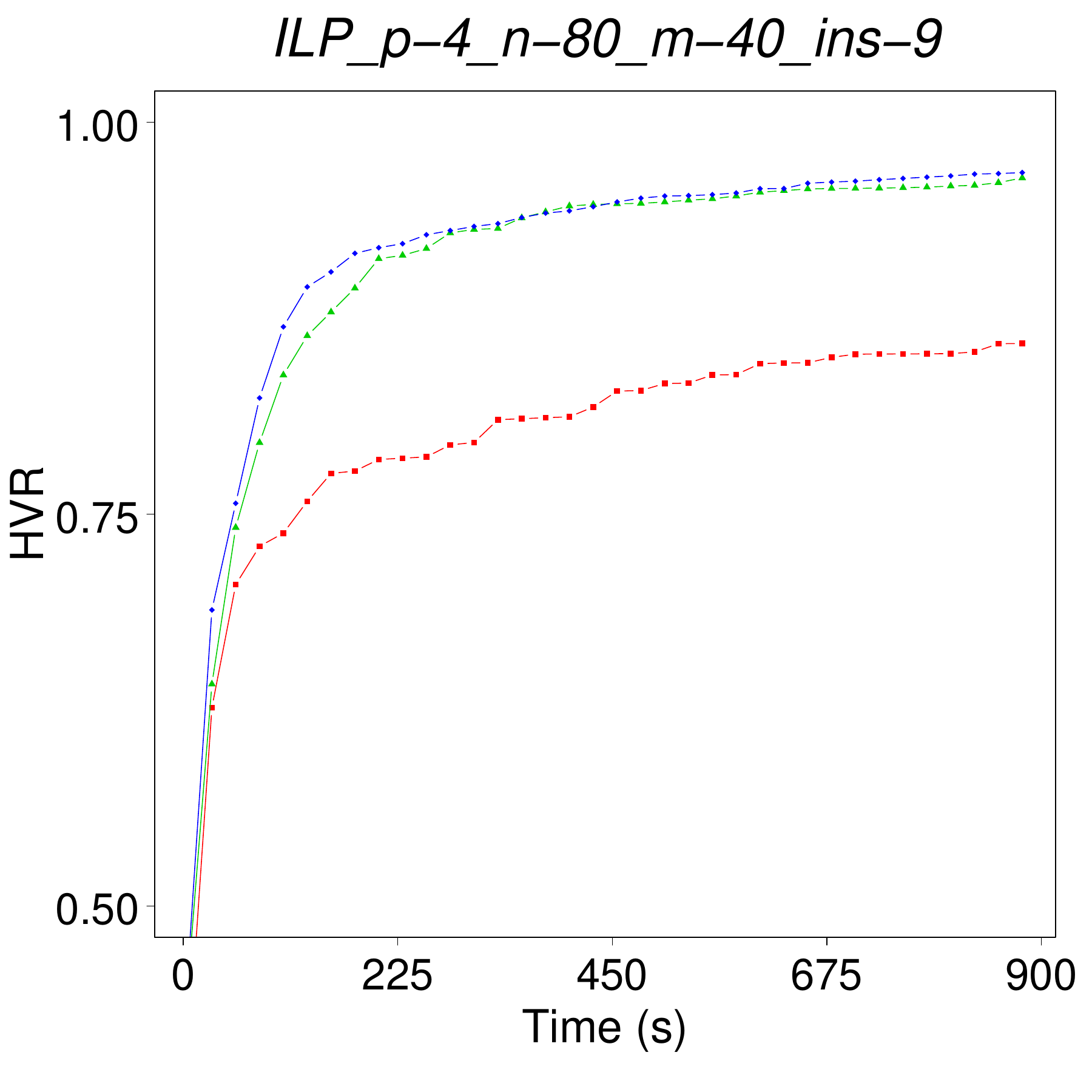}
\end{minipage}
\begin{minipage} {0.326\linewidth}
\includegraphics[scale=0.135]{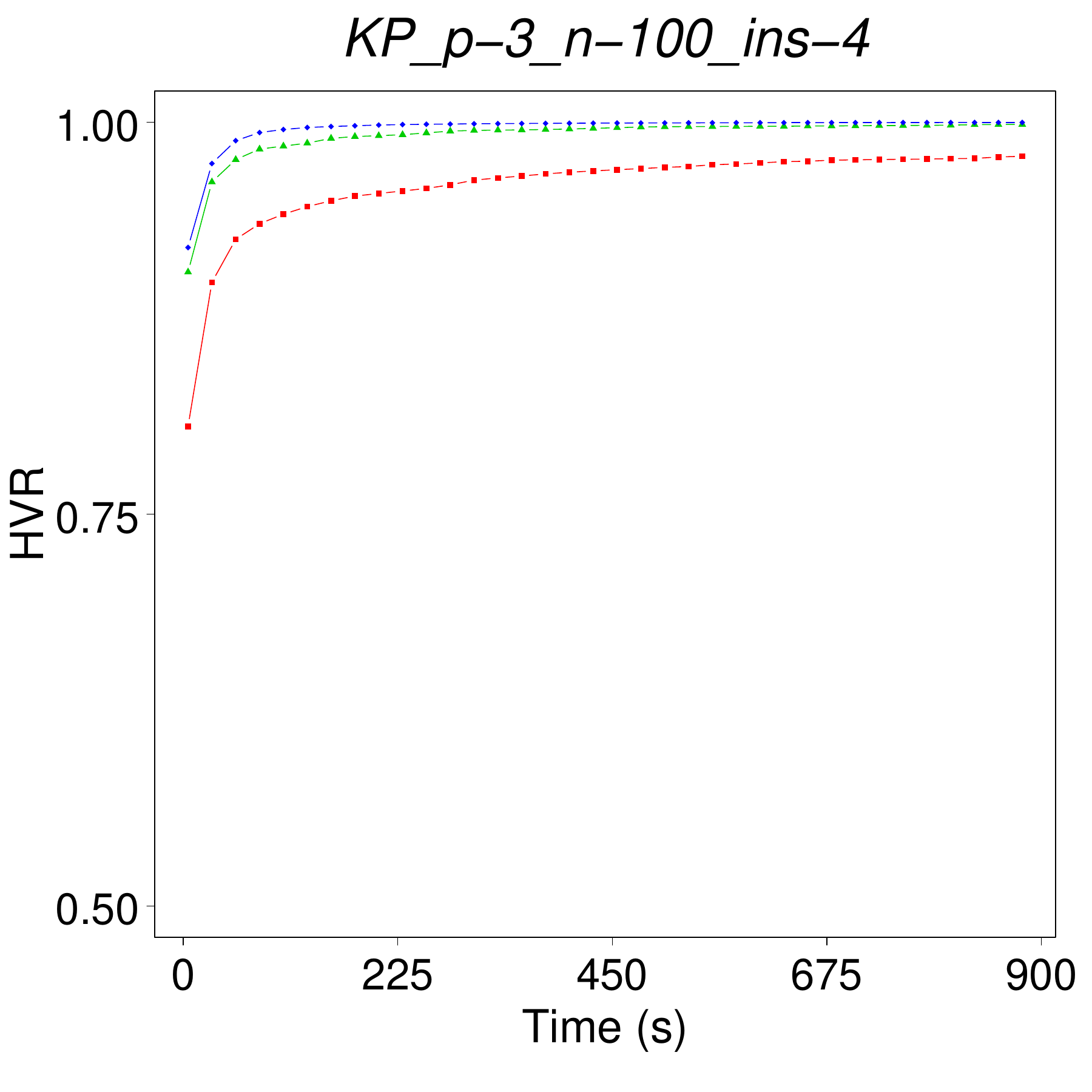}
\end{minipage}

\begin{minipage} {0.5\textwidth}
\includegraphics[scale=0.3]{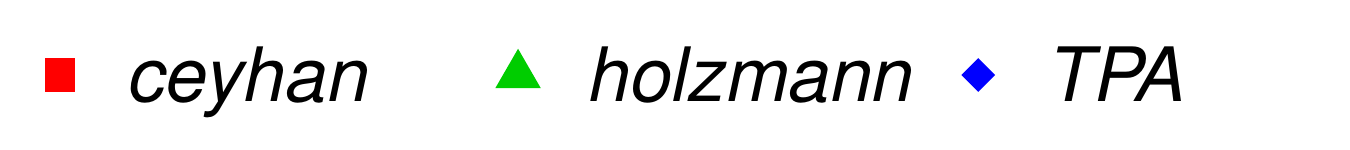} 
\end{minipage}

\end{center}
\caption{ONVGR and HVR as a function of time in 3 selected instances. The higher the value, the better the algorithm.}
\label{fig:hard-1}
\end{figure}

\begin{figure} [h]
\begin{center}
\begin{minipage} {0.326\linewidth}
\includegraphics[scale=0.135]{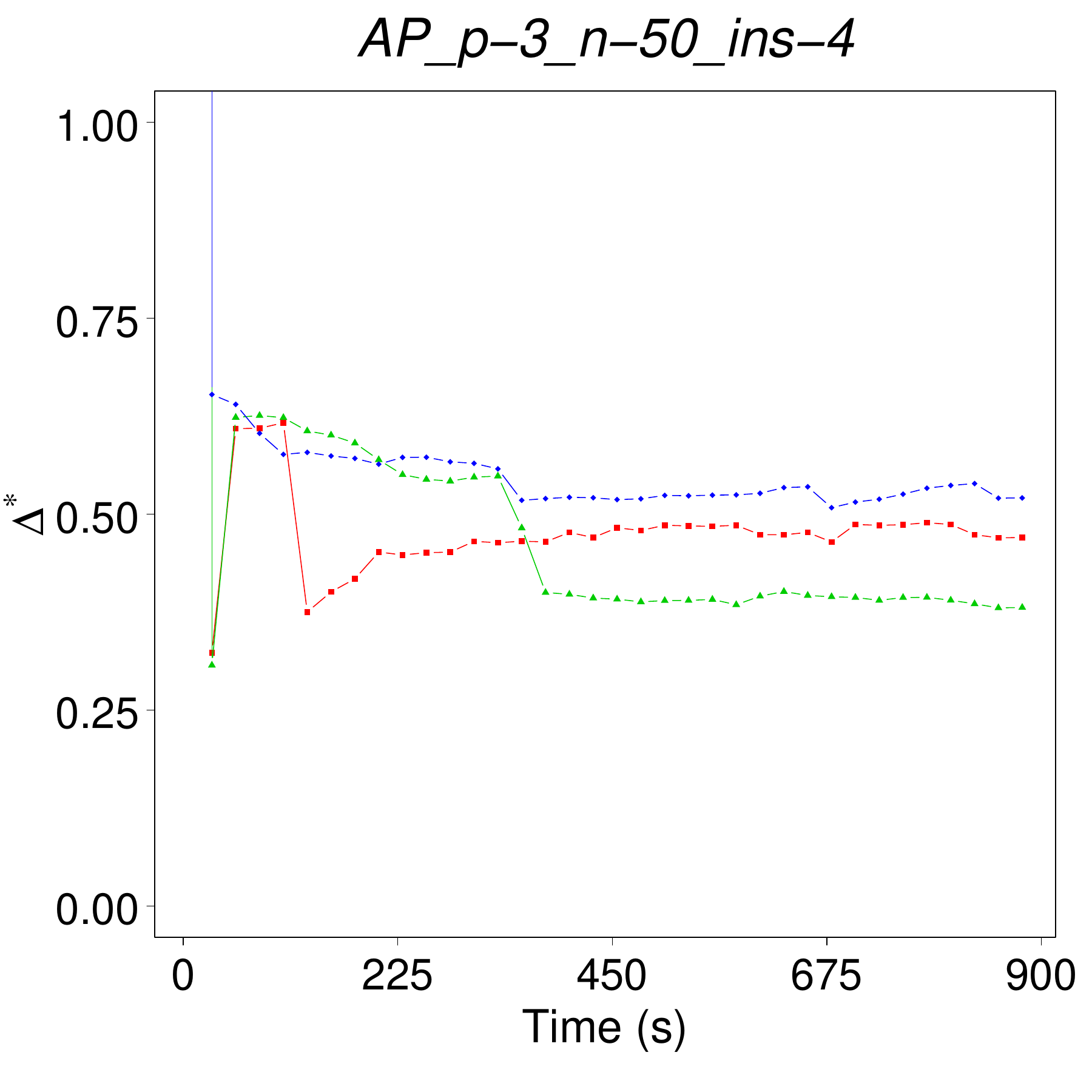}
\end{minipage}
\begin{minipage} {0.326\linewidth}
\includegraphics[scale=0.135]{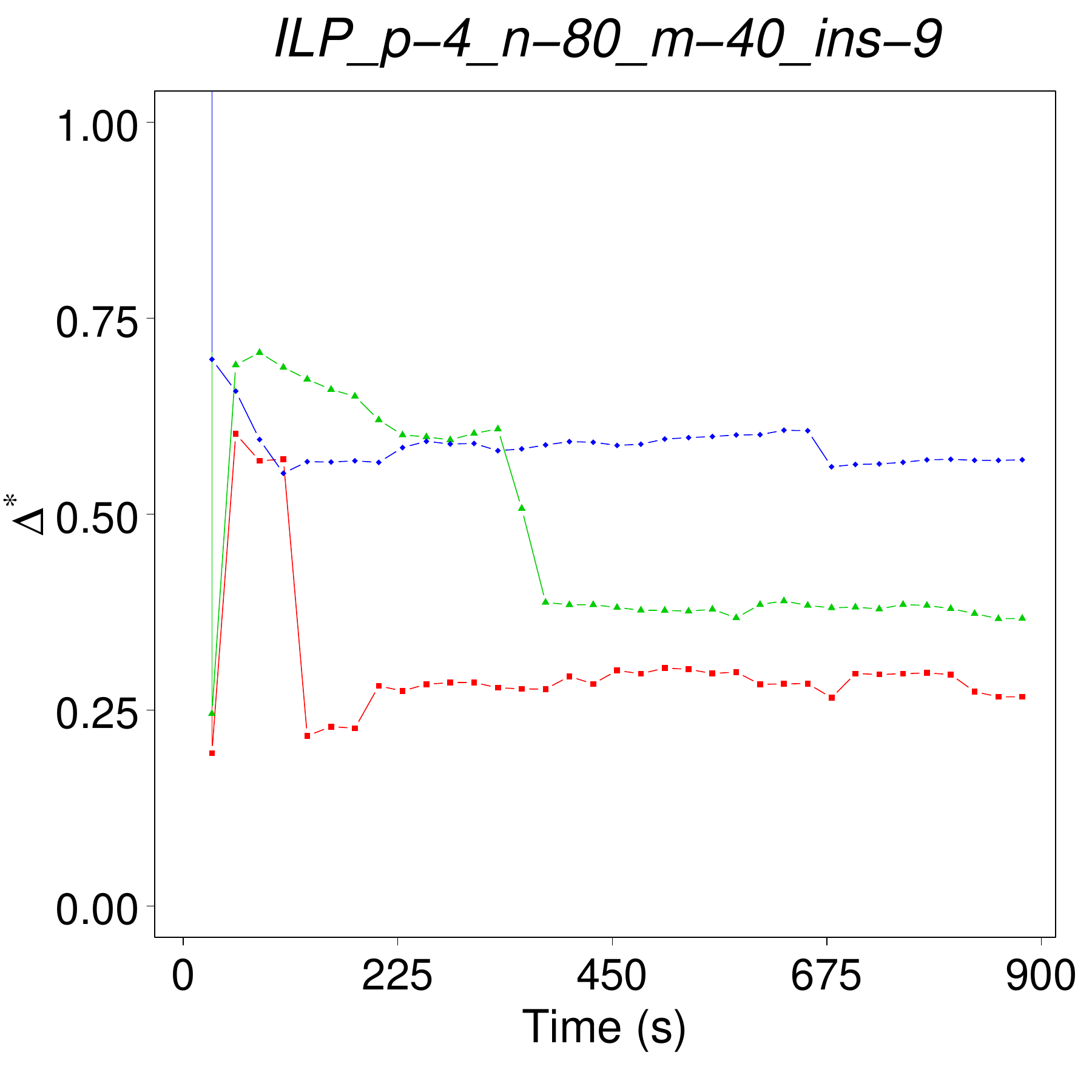}
\end{minipage}
\begin{minipage} {0.326\linewidth}
\includegraphics[scale=0.135]{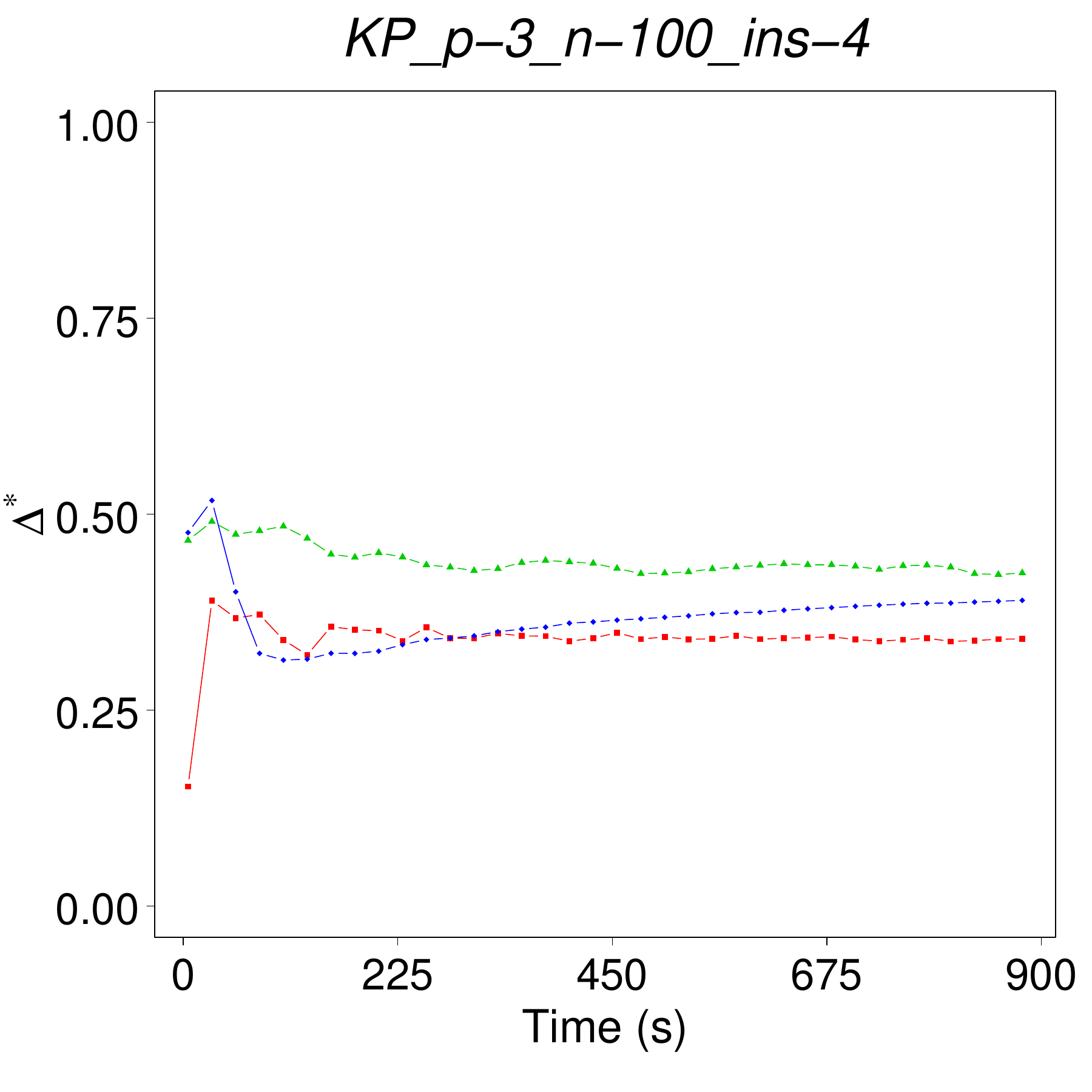}
\end{minipage}

\begin{minipage} {0.326\linewidth}
\includegraphics[scale=0.135]{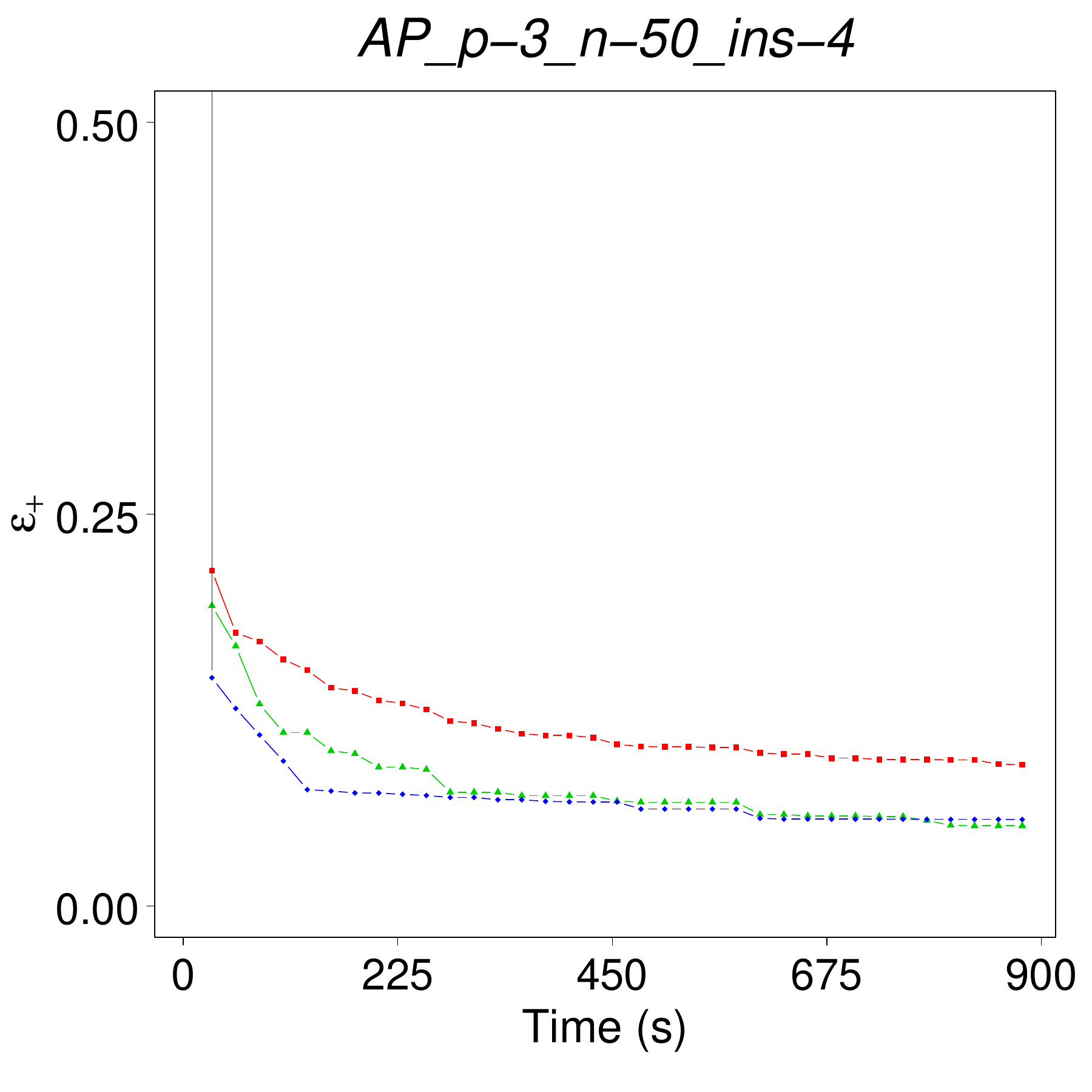}
\end{minipage}
\begin{minipage} {0.326\linewidth}
\includegraphics[scale=0.135]{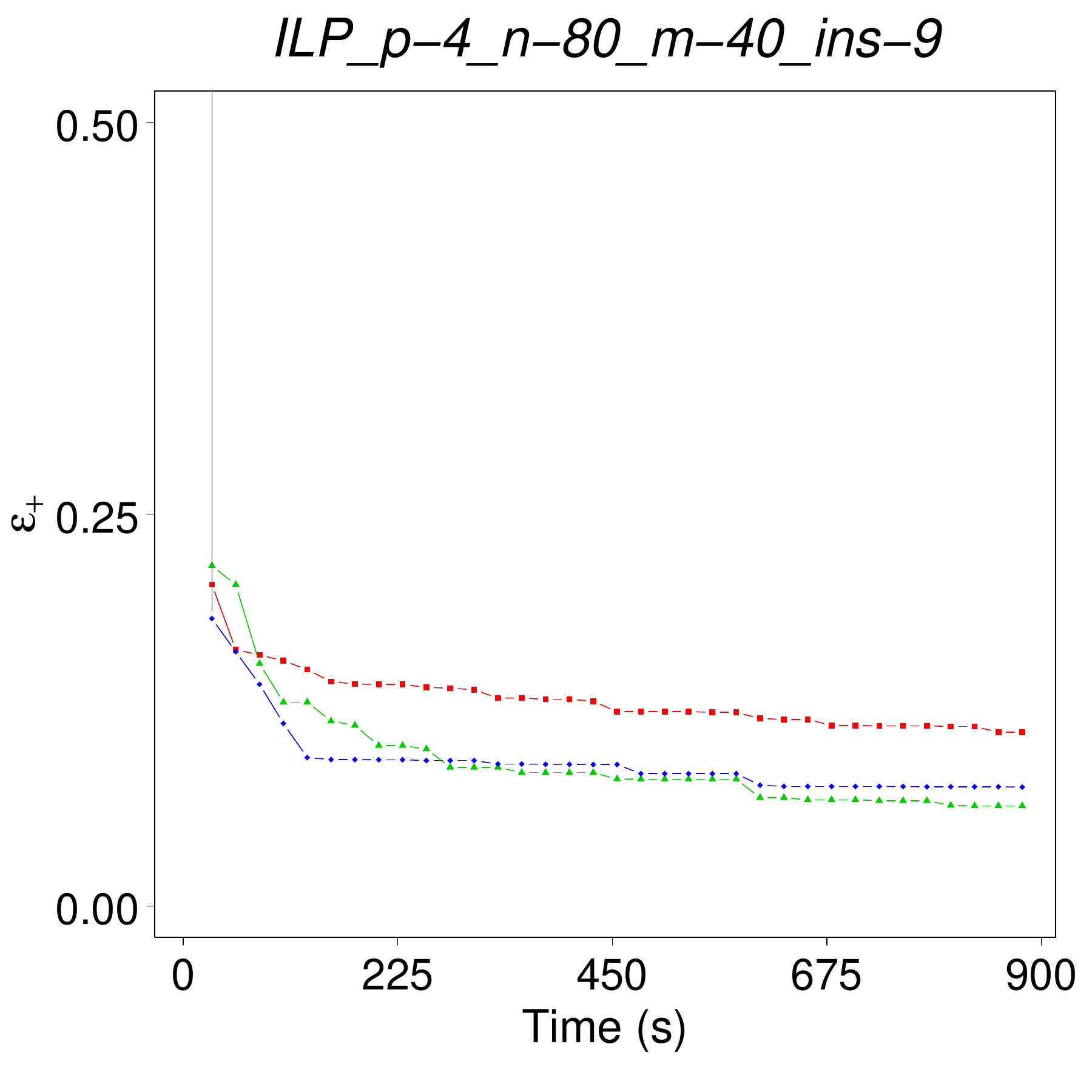}
\end{minipage}
\begin{minipage} {0.326\linewidth}
\includegraphics[scale=0.135]{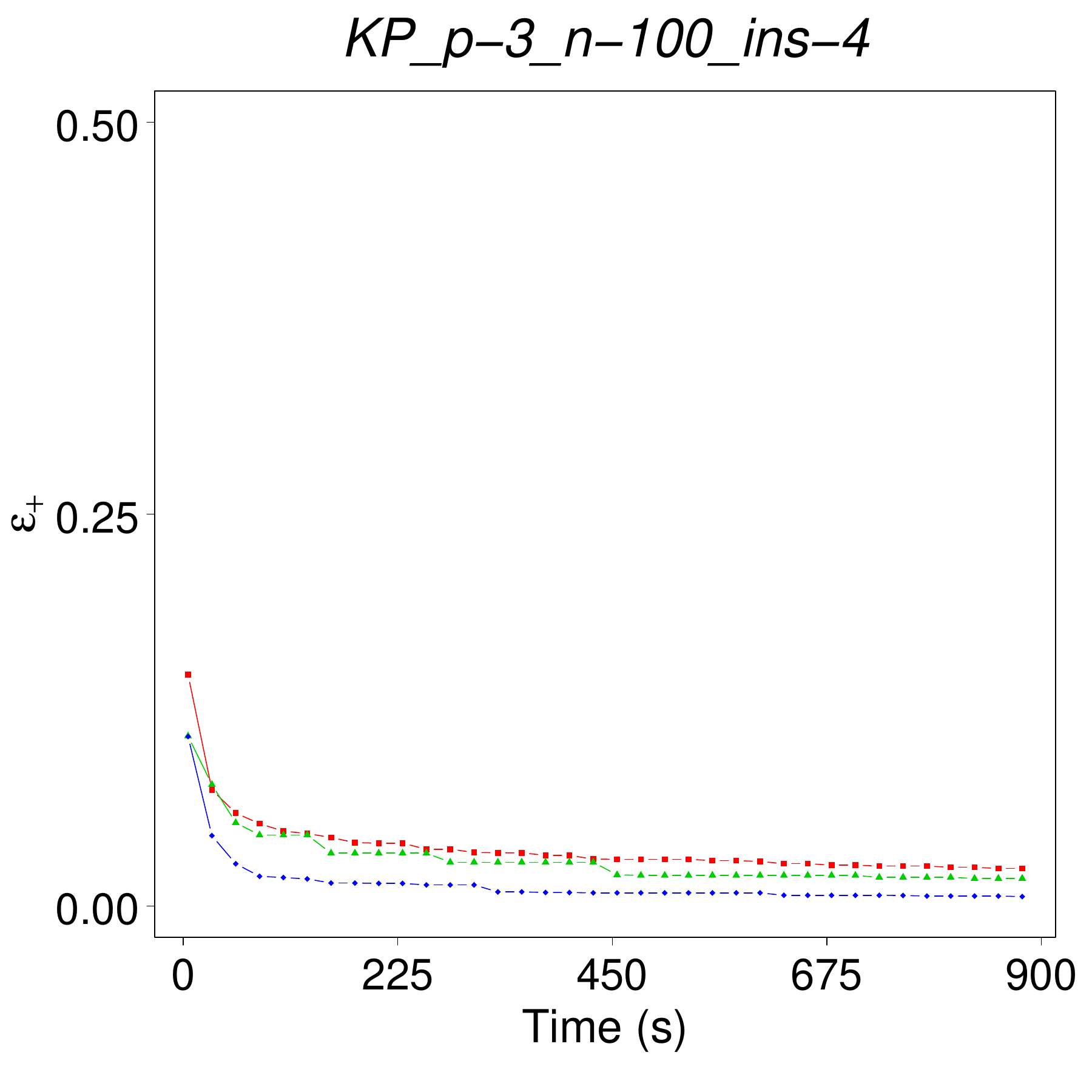}
\end{minipage}

\begin{minipage} {0.5\textwidth}
\includegraphics[scale=0.3]{images/horizontal_legend.pdf} 
\end{minipage}

\end{center}
\caption{$\Delta^{*}$ and $\varepsilon_{+}$ as a function of time in 3 selected instances. The lower the value, the better the algorithm.}
\label{fig:hard-2}
\end{figure}

Figure~\ref{fig:hard-1} shows the behavior of ONVGR and HVR for the three selected instances: \emph{KP\_\,p-3\_\,n-100\_\,ins-4}, which belongs to the multiobjective knapsack problem group, in this case, with 100 variables;  \emph{AP\_\,p-3\_\,n-50\_\,ins-4}, from the assignment problem group, which has 50 agents; and \emph{ILP\_\,p-4\_\,n-80\_\,m-40\_\,ins-9}, from the \emph{ILP} group, which has dimension 4, 80 variables, and 40 constraints. 
Figure~\ref{fig:hard-2} displays the behavior for the other metrics. In the \emph{x}-axis, we show the execution time in seconds. The \emph{y}-axis shows the corresponding quality metric. In this case, the time points are taken every 5 s, up to 900 s. For these particular instances, {\AMOCO} provides the higher number of solutions at any time, HVR is quite similar for {\AMOCO} and \emph{holzmann}, $\Delta^{*}$  is better for \emph{ceyhan} in most of cut-points, and $\varepsilon_{+}$ is similar for the three methods.
 We observe a high increase in HVR during the first minutes and then a stabilization of the value at the end. We see that the evolution of $\Delta^{*}$ in Figure~\ref{fig:hard-2} is not monotonic. This is a consequence of $\Delta^{*}$ not being Pareto compliant.

\subsubsection{Statistical validation}

In order to check if the observed differences are statistically significant or not, we applied the non-parametric Friedman test to compare the three algorithms. We applied the test for each metric and cut-point in the set \{90,180,\ldots,900\}. In the cases in which the $p$-value is below the significance level $\alpha=0.01$, there is a strong evidence that the performances of the algorithms are different. To pinpoint the differences, we did a post hoc analysis using the Nemenyi multiple comparison test. This test compares every pair of algorithms looking for significant differences in a  paired test with the 480 instances, as well as for each class. We chose only the cases in which one method is significantly different than the other two, and then we applied the Wilcoxon signed rank test to confirm that the corresponding algorithm is the best. All these tests are available in the R-package \emph{PMCMRplus}.

A summary with the results of the tests is shown in Table~\ref{tab:posthoc}. We used the initial of the algorithm (C, H or T) in the cases where an algorithm has the best performance with a significance level $\alpha=0.01$ in the three tests (Friedman, Nemenyi and Wilcoxon). We highlight in boldface the cases in which the $p$-value is  below $0.001$. 

The first conclusion is that {\AMOCO} is almost always the best algorithm. The second conclusion is that \emph{Holzmann} is never the best for all the metrics. Another conclusion from the tests is that the behavior of the algorithms is problem-dependent. We observe that, for \emph{ILP} and \emph{KP} problems, \emph{ceyhan} is significantly better with respect to $\Delta^{*}$. In \emph{AP} instances, \emph{ceyhan} is the best in the first 180 s but the differences with {\AMOCO} are not statistically significant. From 270~s to 900~s, \emph{ceyhan} is surpassed by {\AMOCO}. Grouping all the instances, the tests are statistically significant at a significance level of $\alpha = 0.001$ at any cut-point for three of the four metrics, in favor of {\AMOCO}.  
The algorithm proposed in this paper shows a statistically significant difference with \emph{holzmann} for these quality indicators.

\begin{table*}[!h] 
%\tiny
\scriptsize
\centering

\begin{tabular}{ccrrrrrrrrrr}
 \multicolumn{2}{c} {Hypothesis tests} & 90s & 180s & 270s & 360s & 450s & 540s & 630s & 720s & 810s & 900s \tabularnewline
\hline 
\multirow{4}{*} {\emph{AP}} 
& ONVGR & \textsf{\textbf{T}} & \textsf{\textbf{T}} & \textsf{\textbf{T}} & \textsf{\textbf{T}} & \textsf{\textbf{T}} & \textsf{\textbf{T}} & \textsf{\textbf{T}}  & \textsf{\textbf{T}}  & \textsf{\textbf{T}}  & \textsf{\textbf{T}} \\
& HVR   & \textsf{\textbf{T}} & \textsf{\textbf{T}} & \textsf{\textbf{T}} & \textsf{\textbf{T}} & \textsf{\textbf{T}} & \textsf{\textbf{T}} & \textsf{\textbf{T}} & \textsf{\textbf{T}} & \textsf{\textbf{T}} & \textsf{\textbf{T}} \\
& $\Delta^{*}$  &   &  & T & T & \textsf{\textbf{T}} & \textsf{\textbf{T}} & \textsf{\textbf{T}} & \textsf{\textbf{T}} & \textsf{\textbf{T}} & \textsf{\textbf{T}} \\
& $\varepsilon_{+}$  & \textsf{\textbf{T}} & \textsf{\textbf{T}} & \textsf{\textbf{T}} & \textsf{\textbf{T}} & \textsf{\textbf{T}} & \textsf{\textbf{T}} & \textsf{\textbf{T}} & \textsf{\textbf{T}} & \textsf{\textbf{T}} & \textsf{\textbf{T}} \tabularnewline
\hline

\multirow{4}{*} {\emph{ILP}} 
& ONVGR &  &  &  &  &  &  & T  & T  & T  & T \\
& HVR   &  & \textsf{\textbf{T}} & \textsf{\textbf{T}} & \textsf{\textbf{T}} & \textsf{\textbf{T}} & \textsf{\textbf{T}} & \textsf{\textbf{T}} & \textsf{\textbf{T}} & \textsf{\textbf{T}} & \textsf{\textbf{T}} \\
& $\Delta^{*}$  & \textsf{\textbf{C}} & \textsf{\textbf{C}} & \textsf{\textbf{C}} & \textsf{\textbf{C}} & \textsf{\textbf{C}} & \textsf{\textbf{C}} & \textsf{\textbf{C}} & \textsf{\textbf{C}} & \textsf{\textbf{C}} & C \\
& $\varepsilon_{+}$  &  &  &  &  &  & T & T &  & T & T \tabularnewline
\hline

\multirow{4}{*} {\emph{KP}} 
& ONVGR & T & \textsf{\textbf{T}} & T & T & T & T &   &   &   &  \\
& HVR   & T & \textsf{\textbf{T}} & T & \textsf{\textbf{T}} & T & T &  &  &  &  \\
& $\Delta^{*}$  &   & \textsf{\textbf{C}}  & \textsf{\textbf{C}} & C & \textsf{\textbf{C}} & \textsf{\textbf{C}} & \textsf{\textbf{C}} & C &  &  \\
& $\varepsilon_{+}$  & \textsf{\textbf{T}} & \textsf{\textbf{T}} & \textsf{\textbf{T}} & T & T & T &  &  &  &  \tabularnewline
\hline

\multirow{4}{*} {TOTAL} 
& ONVGR & \textsf{\textbf{T}} & \textsf{\textbf{T}} & \textsf{\textbf{T}} & \textsf{\textbf{T}} & \textsf{\textbf{T}} & \textsf{\textbf{T}} & \textsf{\textbf{T}} & \textsf{\textbf{T}} & \textsf{\textbf{T}} & \textsf{\textbf{T}} \\
& HVR & \textsf{\textbf{T}} & \textsf{\textbf{T}} & \textsf{\textbf{T}} & \textsf{\textbf{T}} & \textsf{\textbf{T}} & \textsf{\textbf{T}} & \textsf{\textbf{T}} & \textsf{\textbf{T}} & \textsf{\textbf{T}} & \textsf{\textbf{T}} \\
& $\Delta^{*}$  & \textsf{\textbf{C}}  &  &  &  &  &  &  &  &  &  \\
& $\varepsilon_{+}$ & \textsf{\textbf{T}} & \textsf{\textbf{T}} & \textsf{\textbf{T}} & \textsf{\textbf{T}} & \textsf{\textbf{T}} & \textsf{\textbf{T}} & \textsf{\textbf{T}} & \textsf{\textbf{T}} & \textsf{\textbf{T}} & \textsf{\textbf{T}} \tabularnewline
\hline

\end{tabular}

\caption{Statistical tests for each class and metric at different time points. The initial of the best algorithm is shown when the differences with the others are statistically significant at level $\alpha=0.01$. If the differences are also significant at level $\alpha=0.001$, the initial of the algorithm is marked in bold.}

\label{tab:posthoc}
\end{table*}

\section{Conclusions and future work}
We have designed a new anytime algorithm that allows the Pareto front to be calculated in multiobjective combinatorial optimization problems. The front is well-spread at any time and has good values for the metrics ONVGR, HV, $\Delta^{*}$, and $\varepsilon_{+}$. We have first carried out an exhaustive study of the literature, and have identified two state-of-the-art methods which produce a well-spread set of non-dominated points at any time and are not outperformed by other methods. They were proposed by Ceyhan et al.~\cite{ceyhan2019finding} and Holzmann and Smith~\cite{holzmann2018solving}.

The algorithm we propose, called {\AMOCO}, is based on an existing framework to solve MOCO problems~\cite{dachert2015linear}, which has been adapted to design an effective anytime algorithm. The new contributions are: the establishment of a new strategy to select the appropriate search space region as the next box to explore, a new way of partitioning the search space after finding a new non-dominated point, and the definition of a new quality function to set the priority for the new regions to explore. These three new contributions have a positive influence in the spread of the solutions.

We compared {\AMOCO} with \emph{ceyhan} and \emph{holzmann} in several ways. A deep performance analysis was done using 480 instances and executing the algorithms 30 times in each of them. We compared the algorithms using four different metrics well-known in the MOO literature. Finally, we have statistically checked if the observed differences are statistically significant using Friedman test, Nemenyi multiple comparison test, and Wilcoxon signed rank test. 
We can statistically confirm (for this benchmark of instances) that {\AMOCO} is better than the state-of-the-art methods in almost all the cases, improving the number of solutions, hypervolume and additive epsilon indicator for all the time cut-points used, which indicates a better anytime behavior. In the case of the general spread (which is not Pareto compliant), \emph{ceyhan} is better in the first 90~s, and there is no conclusion for the rest of cut-points when considering all the instances. 

Future work includes the extension of the experimental study to other benchmarks, to analyze in which kind of problems our proposed algorithm works best. We can also apply {\AMOCO} to multi-objective industrial problems, where a few efficient solutions covering the objective space are required by decision makers in a short time. The ideas introduced in this paper can also be applied to heuristic algorithms, in the field of evolutionary multi-objective optimization, for example, to improve the performance of multi-objective meta-heuristics. We can also combine heuristics with exact methods to create new hybrids enjoying the advantages of both fields.

\section*{Acknowledgments}
This research has been partially funded by the Spanish Ministry of Economy and Competitiveness (MINECO) and the European Regional Development Fund (FEDER) under contract  TIN2017-88213-R (6city project), 
the European Research Council under contract H2020-ICT-2019-3 (TAILOR project), 
the University of M\'alaga, Consejer\'ia de Econom\'ia y Conocimiento de la Junta de Andaluc\'ia and FEDER under contract UMA18-FEDERJA-003 (PRECOG project), 
the Ministry of Science, Innovation and Universities and FEDER under contract RTC-2017-6714-5,
and the University of Málaga under contract PPIT.UMA.B1.2017/07 (EXHAURO Project).

%\bibliographystyle{amsplain}
%\bibliographystyle{plainnat}
%\bibliography{MO_bibliography.bib}

\begin{thebibliography}{39}
\providecommand{\natexlab}[1]{#1}
\providecommand{\url}[1]{\texttt{#1}}
\expandafter\ifx\csname urlstyle\endcsname\relax
  \providecommand{\doi}[1]{doi: #1}\else
  \providecommand{\doi}{doi: \begingroup \urlstyle{rm}\Url}\fi

\bibitem[Ceyhan et~al.(2019)Ceyhan, K{\"o}ksalan, and
  Lokman]{ceyhan2019finding}
G.~Ceyhan, M.~K{\"o}ksalan, and B.~Lokman.
\newblock Finding a representative nondominated set for multi-objective mixed
  integer programs.
\newblock \emph{European Journal of Operational Research}, 272\penalty0
  (1):\penalty0 61--77, 2019.

\bibitem[D{\"a}chert and Klamroth(2015)]{dachert2015linear}
K.~D{\"a}chert and K.~Klamroth.
\newblock A linear bound on the number of scalarizations needed to solve
  discrete tricriteria optimization problems.
\newblock \emph{Journal of Global Optimization}, 61\penalty0 (4):\penalty0
  643--676, 2015.

\bibitem[D{\"a}chert et~al.(2017)D{\"a}chert, Klamroth, Lacour, and
  Vanderpooten]{dachert2017efficient}
K.~D{\"a}chert, K.~Klamroth, R.~Lacour, and D.~Vanderpooten.
\newblock Efficient computation of the search region in multi-objective
  optimization.
\newblock \emph{European Journal of Operational Research}, 260\penalty0
  (3):\penalty0 841--855, 2017.

\bibitem[Dean and Boddy(1988)]{dean1988solving}
T.~Dean and M.~Boddy.
\newblock Solving time-dependent planning problems.
\newblock In \emph{Proceedings of the 7th National Conference on Artificial
  Intelligence}, pages 49--54, 1988.

\bibitem[Deb et~al.(2000)Deb, Agrawal, Pratap, and Meyarivan]{deb2000fast}
K.~Deb, S.~Agrawal, A.~Pratap, and T.~Meyarivan.
\newblock A fast elitist non-dominated sorting genetic algorithm for
  multi-objective optimization: {NSGA-II}.
\newblock In \emph{International Conference on Parallel Problem Solving from
  Nature}, pages 849--858. Springer, 2000.

\bibitem[Dhaenens et~al.(2010)Dhaenens, Lemesre, and Talbi]{dhaenens2010k}
C.~Dhaenens, J.~Lemesre, and E.~Talbi.
\newblock {K-PPM}: A new exact method to solve multi-objective combinatorial
  optimization problems.
\newblock \emph{European Journal of Operational Research}, 200\penalty0
  (1):\penalty0 45--53, 2010.

\bibitem[Ehrgott(2005)]{ehrgott2005multicriteria}
M.~Ehrgott.
\newblock \emph{Multicriteria optimization}, volume 491.
\newblock Springer Science \& Business Media, 2005.

\bibitem[Ehrgott and Tenfelde-Podehl(2003)]{ehrgott2003computation}
M.~Ehrgott and D.~Tenfelde-Podehl.
\newblock Computation of ideal and nadir values and implications for their use
  in {MCDM} methods.
\newblock \emph{European Journal of Operational Research}, 151\penalty0
  (1):\penalty0 119--139, 2003.

\bibitem[Fonseca et~al.(2006)Fonseca, Paquete, and
  L{\'o}pez-Ib{\'a}\~nez]{fonseca2006improved}
C.M. Fonseca, L.~Paquete, and M.~L{\'o}pez-Ib{\'a}\~nez.
\newblock An improved dimension-sweep algorithm for the hypervolume indicator.
\newblock In \emph{Proceedings of the IEEE Congress on Evolutionary
  Computation}, pages 1157--1163. IEEE, 2006.

\bibitem[Gary and Johnson(1979)]{gary1979computers}
M.R. Gary and D.S. Johnson.
\newblock Computers and intractability: A guide to the theory of
  {NP}-completeness, 1979.

\bibitem[Holzmann and Smith(2018)]{holzmann2018solving}
T.~Holzmann and J.C. Smith.
\newblock Solving discrete multi-objective optimization problems using modified
  augmented weighted {T}chebychev scalarizations.
\newblock \emph{European Journal of Operational Research}, 271\penalty0
  (2):\penalty0 436--449, 2018.

\bibitem[Hutson(2018)]{Hutson725}
M.~Hutson.
\newblock Artificial intelligence faces reproducibility crisis.
\newblock \emph{Science}, 359\penalty0 (6377):\penalty0 725--726, 2018.

\bibitem[Jiang et~al.(2014)Jiang, Ong, Zhang, and Feng]{jiang2014consistencies}
S.~Jiang, Y.~Ong, J.~Zhang, and L.~Feng.
\newblock Consistencies and contradictions of performance metrics in
  multiobjective optimization.
\newblock \emph{IEEE Transactions on Cybernetics}, 44\penalty0 (12):\penalty0
  2391--2404, 2014.

\bibitem[Kirlik and Say{\i}n(2014)]{kirlik2014new}
G.~Kirlik and S.~Say{\i}n.
\newblock A new algorithm for generating all nondominated solutions of
  multiobjective discrete optimization problems.
\newblock \emph{European Journal of Operational Research}, 232\penalty0
  (3):\penalty0 479--488, 2014.

\bibitem[Klamroth et~al.(2015)Klamroth, Lacour, and
  Vanderpooten]{klamroth2015representation}
K.~Klamroth, R.~Lacour, and D.~Vanderpooten.
\newblock On the representation of the search region in multi-objective
  optimization.
\newblock \emph{European Journal of Operational Research}, 245\penalty0
  (3):\penalty0 767--778, 2015.

\bibitem[Klein and Hannan(1982)]{klein1982algorithm}
D.~Klein and E.~Hannan.
\newblock An algorithm for the multiple objective integer linear programming
  problem.
\newblock \emph{European Journal of Operational Research}, 9\penalty0
  (4):\penalty0 378--385, 1982.

\bibitem[Laumanns et~al.(2005)Laumanns, Thiele, and
  Zitzler]{laumanns2005adaptive}
M.~Laumanns, L.~Thiele, and E.~Zitzler.
\newblock An adaptive scheme to generate the {P}areto front based on the
  epsilon-constraint method.
\newblock In \emph{Dagstuhl Seminar Proceedings}. Schloss
  Dagstuhl-Leibniz-Zentrum f{\"u}r Informatik, 2005.

\bibitem[Laumanns et~al.(2006)Laumanns, Thiele, and
  Zitzler]{laumanns2006efficient}
M.~Laumanns, L.~Thiele, and E.~Zitzler.
\newblock An efficient, adaptive parameter variation scheme for metaheuristics
  based on the epsilon-constraint method.
\newblock \emph{European Journal of Operational Research}, 169\penalty0
  (3):\penalty0 932--942, 2006.

\bibitem[Lemesre et~al.(2007)Lemesre, Dhaenens, and Talbi]{lemesre2007parallel}
J.~Lemesre, C.~Dhaenens, and E.~Talbi.
\newblock Parallel partitioning method ({PPM}): A new exact method to solve
  bi-objective problems.
\newblock \emph{Computers \& Operations Research}, 34\penalty0 (8):\penalty0
  2450--2462, 2007.

\bibitem[Li and Yao(2019)]{li2019quality}
M.~Li and X.~Yao.
\newblock Quality evaluation of solution sets in multiobjective optimisation: A
  survey.
\newblock \emph{ACM Computing Surveys}, 52\penalty0 (2):\penalty0 1--38, 2019.

\bibitem[Liefooghe and Derbel(2016)]{liefooghe2016correlation}
A.~Liefooghe and B.~Derbel.
\newblock A correlation analysis of set quality indicator values in
  multiobjective optimization.
\newblock In \emph{Proceedings of the Genetic and Evolutionary Computation
  Conference 2016}, pages 581--588, 2016.

\bibitem[Lokman and K{\"o}ksalan(2013)]{lokman2013finding}
B.~Lokman and M.~K{\"o}ksalan.
\newblock Finding all nondominated points of multi-objective integer programs.
\newblock \emph{Journal of Global Optimization}, 57\penalty0 (2):\penalty0
  347--365, 2013.

\bibitem[L\'opez-Ib\'a\~nez and St{\"u}tzle(2014)]{lopez2014automatically}
M.~L\'opez-Ib\'a\~nez and T.~St{\"u}tzle.
\newblock Automatically improving the anytime behaviour of optimisation
  algorithms.
\newblock \emph{European Journal of Operational Research}, 235\penalty0
  (3):\penalty0 569--582, 2014.

\bibitem[Masin and Bukchin(2008)]{masin2008diversity}
M.~Masin and Y.~Bukchin.
\newblock Diversity maximization approach for multiobjective optimization.
\newblock \emph{Operations Research}, 56\penalty0 (2):\penalty0 411--424, 2008.

\bibitem[{\"O}zlen and Azizo{\u{g}}lu(2009)]{ozlen2009multi}
M.~{\"O}zlen and M.~Azizo{\u{g}}lu.
\newblock Multi-objective integer programming: a general approach for
  generating all non-dominated solutions.
\newblock \emph{European Journal of Operational Research}, 199\penalty0
  (1):\penalty0 25--35, 2009.

\bibitem[Ozlen et~al.(2014)Ozlen, Burton, and MacRae]{ozlen2014multi}
M.~Ozlen, B.A. Burton, and C.A.G. MacRae.
\newblock Multi-objective integer programming: An improved recursive algorithm.
\newblock \emph{Journal of Optimization Theory and Applications}, 160\penalty0
  (2):\penalty0 470--482, 2014.

\bibitem[{\"O}zpeynirci and K{\"o}ksalan(2010)]{ozpeynirci2010exact}
{\"O}.~{\"O}zpeynirci and M.~K{\"o}ksalan.
\newblock An exact algorithm for finding extreme supported nondominated points
  of multiobjective mixed integer programs.
\newblock \emph{Management Science}, 56\penalty0 (12):\penalty0 2302--2315,
  2010.

\bibitem[Przybylski et~al.(2010{\natexlab{a}})Przybylski, Gandibleux, and
  Ehrgott]{przybylski2010recursive}
A.~Przybylski, X.~Gandibleux, and M.~Ehrgott.
\newblock A recursive algorithm for finding all nondominated extreme points in
  the outcome set of a multiobjective integer programme.
\newblock \emph{INFORMS Journal on Computing}, 22\penalty0 (3):\penalty0
  371--386, 2010{\natexlab{a}}.

\bibitem[Przybylski et~al.(2010{\natexlab{b}})Przybylski, Gandibleux, and
  Ehrgott]{przybylski2010two}
A.~Przybylski, X.~Gandibleux, and M.~Ehrgott.
\newblock A two phase method for multi-objective integer programming and its
  application to the assignment problem with three objectives.
\newblock \emph{Discrete Optimization}, 7\penalty0 (3):\penalty0 149--165,
  2010{\natexlab{b}}.

\bibitem[Rostami and Neri(2017)]{rostami2017fast}
S.~Rostami and F.~Neri.
\newblock A fast hypervolume driven selection mechanism for many-objective
  optimisation problems.
\newblock \emph{Swarm and Evolutionary Computation}, 34:\penalty0 50--67, 2017.

\bibitem[Rostami et~al.(2020)Rostami, Neri, and
  Gyaurski]{rostami2020algorithmic}
S.~Rostami, F.~Neri, and K.~Gyaurski.
\newblock On algorithmic descriptions and software implementations for
  multi-objective optimisation: A comparative study.
\newblock \emph{SN Computer Science}, 1\penalty0 (5):\penalty0 1--23, 2020.

\bibitem[Sylva and Crema(2004)]{sylva2004method}
J.~Sylva and A.~Crema.
\newblock A method for finding the set of non-dominated vectors for multiple
  objective integer linear programs.
\newblock \emph{European Journal of Operational Research}, 158\penalty0
  (1):\penalty0 46--55, 2004.

\bibitem[Sylva and Crema(2007)]{sylva2007method}
J.~Sylva and A.~Crema.
\newblock A method for finding well-dispersed subsets of non-dominated vectors
  for multiple objective mixed integer linear programs.
\newblock \emph{European Journal of Operational Research}, 180\penalty0
  (3):\penalty0 1011--1027, 2007.

\bibitem[Tenfelde-Podehl(2003)]{tenfelde2003recursive}
D.~Tenfelde-Podehl.
\newblock \emph{A recursive algorithm for multiobjective combinatorial
  optimization problems with q criteria}.
\newblock Universit{\"a}t Graz/Technische Universit{\"a}t Graz. SFB
  F003-Optimierung und Kontrolle, 2003.

\bibitem[Ulungu and Teghem(1995)]{ulungu1995two}
E.L. Ulungu and J.~Teghem.
\newblock The two phases method: An efficient procedure to solve bi-objective
  combinatorial optimization problems.
\newblock \emph{Foundations of Computing and Decision Sciences}, 20\penalty0
  (2):\penalty0 149--165, 1995.

\bibitem[While et~al.(2012)While, Bradstreet, and Barone]{while2012fast}
L.~While, L.~Bradstreet, and L.~Barone.
\newblock A fast way of calculating exact hypervolumes.
\newblock \emph{IEEE Transactions on Evolutionary Computation}, 16\penalty0
  (1):\penalty0 86--95, 2012.

\bibitem[Zhou et~al.(2006)Zhou, Jin, Zhang, Sendhoff, and
  Tsang]{zhou2006combining}
A.~Zhou, Y.~Jin, Q.~Zhang, B.~Sendhoff, and E.~Tsang.
\newblock Combining model-based and genetics-based offspring generation for
  multi-objective optimization using a convergence criterion.
\newblock In \emph{2006 IEEE International Conference on Evolutionary
  Computation}, pages 892--899. IEEE, 2006.

\bibitem[Zitzler et~al.(2003)Zitzler, Thiele, Laumanns, Fonseca, and
  Da~Fonseca]{zitzler2003performance}
E.~Zitzler, L.~Thiele, M.~Laumanns, C.M. Fonseca, and V.G. Da~Fonseca.
\newblock Performance assessment of multiobjective optimizers: An analysis and
  review.
\newblock \emph{IEEE Transactions on Evolutionary Computation}, 7\penalty0
  (2):\penalty0 117--132, 2003.

\bibitem[Zitzler et~al.(2007)Zitzler, Brockhoff, and
  Thiele]{zitzler2007hypervolume}
E.~Zitzler, D.~Brockhoff, and L.~Thiele.
\newblock The hypervolume indicator revisited: On the design of
  {P}areto-compliant indicators via weighted integration.
\newblock In \emph{International Conference on Evolutionary Multi-Criterion
  Optimization}, pages 862--876. Springer, 2007.

\end{thebibliography}

\end{document}